\documentclass[sigconf]{acmart}

\usepackage{pifont}
\usepackage{multirow}
\usepackage{makecell}
\usepackage{colortbl}
\usepackage{arydshln}
\usepackage{tabularx}

\AtBeginDocument{%
  }

\copyrightyear{2025}
\acmYear{2025}
\setcopyright{acmlicensed}\acmConference[MM '25]{Proceedings of the 33rd ACM International Conference on Multimedia}{October 27--31, 2025}{Dublin, Ireland}
\acmBooktitle{Proceedings of the 33rd ACM International Conference on Multimedia (MM '25), October 27--31, 2025, Dublin, Ireland}
\acmDOI{10.1145/3746027.3755332}
\acmISBN{979-8-4007-2035-2/2025/10}

\begin{document}

%%
%% The "title" command has an optional parameter,
%% allowing the author to define a "short title" to be used in page headers.
\title{FVQ: A Large-Scale Dataset and an LMM-based Method for Face Video Quality Assessment}

%%
%% The "author" command and its associated commands are used to define
%% the authors and their affiliations.
%% Of note is the shared affiliation of the first two authors, and the
%% "authornote" and "authornotemark" commands
%% used to denote shared contribution to the research.
\author{Sijing Wu}
\affiliation{%
  \institution{Shanghai Jiao Tong University}
  \city{Shanghai}
  \country{China}
}
\email{wusijing@sjtu.edu.cn}
\orcid{0009-0000-7753-1596}

\author{Yunhao Li}
\affiliation{%
  \institution{Shanghai Jiao Tong University}
  \city{Shanghai}
  \country{China}}
\email{lyhsjtu@sjtu.edu.cn}

\author{Ziwen Xu}
\affiliation{%
  \institution{Shanghai Jiao Tong University}
  \city{Shanghai}
  \country{China}}
\email{southbay@sjtu.edu.cn}

\author{Yixuan Gao}
\affiliation{%
  \institution{Shanghai Jiao Tong University}
  \city{Shanghai}
  \country{China}}
\email{gaoyixuan@sjtu.edu.cn}

\author{Huiyu Duan}
\authornote{Corresponding author.}
\affiliation{%
  \institution{Shanghai Jiao Tong University}
  \city{Shanghai}
  \country{China}}
\email{huiyuduan@sjtu.edu.cn}

\author{Wei Sun}
\affiliation{%
  \institution{East China Normal University}
  \city{Shanghai}
  \country{China}}
\email{sunguwei@sjtu.edu.cn}

\author{Guangtao Zhai}
\authornotemark[1]
\affiliation{%
  \institution{Shanghai Jiao Tong University}
  \city{Shanghai}
  \country{China}}
\email{zhaiguangtao@sjtu.edu.cn}

%%
%% By default, the full list of authors will be used in the page
%% headers. Often, this list is too long, and will overlap
%% other information printed in the page headers. This command allows
%% the author to define a more concise list
%% of authors' names for this purpose.
\renewcommand{\shortauthors}{Sijing Wu et al.}

%%
%% The abstract is a short summary of the work to be presented in the
%% article.
\begin{abstract}
Face video quality assessment (FVQA) deserves to be explored in addition to general video quality assessment (VQA), as face videos are the primary content on social media platforms and human visual system (HVS) is particularly sensitive to human faces.
However, FVQA is rarely explored due to the lack of large-scale FVQA datasets.
To fill this gap, we present the first large-scale in-the-wild FVQA dataset, \textbf{FVQ-20K}, which contains 20,000 in-the-wild face videos together with corresponding mean opinion score (MOS) annotations.
Along with the FVQ-20K dataset, we further propose a specialized FVQA method named \textbf{FVQ-Rater} to achieve human-like rating and scoring for face video, which is the first attempt to explore the potential of large multimodal models (LMMs) for the FVQA task.
Concretely, we elaborately extract multi-dimensional features including spatial features, temporal features, and face-specific features (\textit{i.e.}, portrait features and face embeddings) to provide comprehensive visual information, and take advantage of the LoRA-based instruction tuning technique to achieve quality-specific fine-tuning, which shows superior performance on both FVQ-20K and CFVQA datasets.
Extensive experiments and comprehensive analysis demonstrate the significant potential of the FVQ-20K dataset and FVQ-Rater method in promoting the development of FVQA.
The code and dataset will be released at: \url{https://github.com/wsj-sjtu/FVQ}.
\end{abstract}

%%
%% The code below is generated by the tool at http://dl.acm.org/ccs.cfm.
%% Please copy and paste the code instead of the example below.
%%


%%
%% Keywords. The author(s) should pick words that accurately describe
%% the work being presented. Separate the keywords with commas.
%% A "teaser" image appears between the author and affiliation
%% information and the body of the document, and typically spans the
%% page.
\begin{teaserfigure}
  \vspace{-2mm}
  \includegraphics[width=\textwidth]{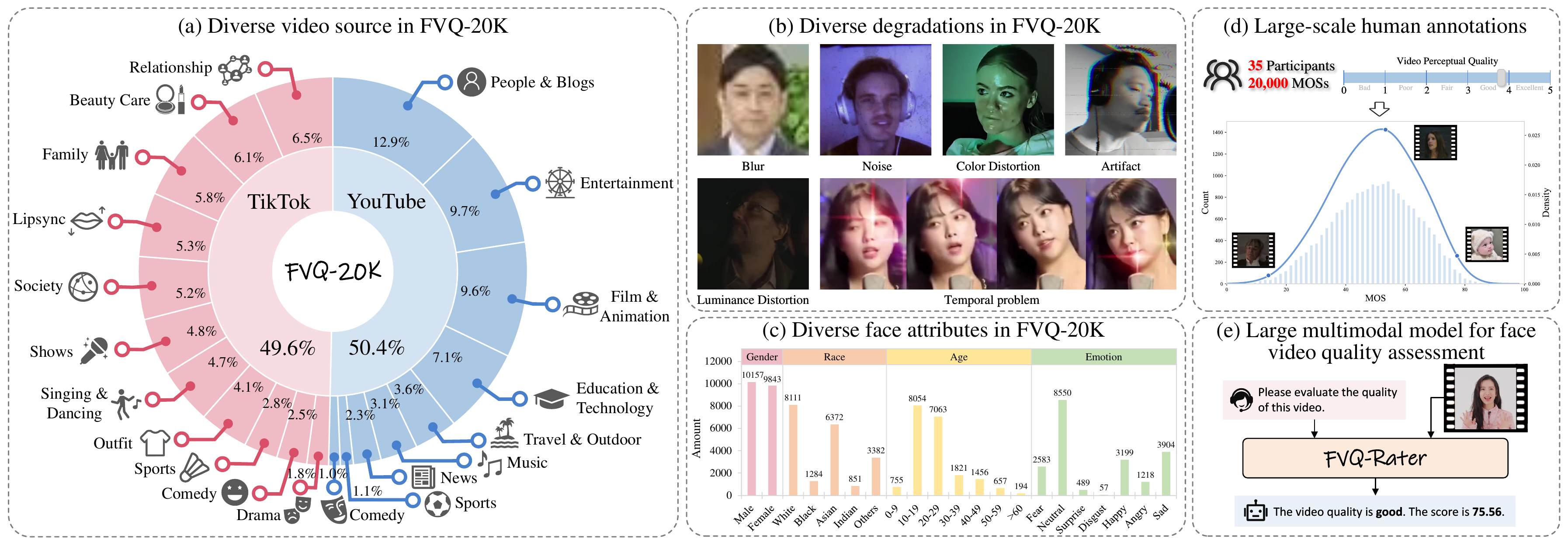}
  \vspace{-4mm}
  \caption{We explore the in-the-wild FVQA problem for the first time. Concretely, we present \textbf{FVQ-20K}, the first large-scale in-the-wild FVQA dataset, which contains 20,000 face videos with (a) diverse source video content, (b) various distortions in both spatial and temporal domains, (c) a variety of facial attributes, and (d) high-quality MOS annotation for each video. Along with the FVQ-20K dataset, we propose (e) \textbf{FVQ-Rater}, the first LMM-based method elaborately designed for the FVQA task.}
  \label{fig:teaser}
\end{teaserfigure}

% \received{20 February 2007}
% \received[revised]{12 March 2009}
% \received[accepted]{5 June 2009}

%%
%% This command processes the author and affiliation and title
%% information and builds the first part of the formatted document.
\maketitle

\begin{table*}
% \vspace{-8mm}
\centering
\caption{Summary and comparison of popular VQA, FIQA, and FVQA datasets.}
\vspace{-2mm}

% \vspace{-0.5em}
\resizebox{\textwidth}{!}{
\begin{tabular}{lcccccccccc}
\toprule

Dataset	&	Year	&	Task	&	Type	&	Total Data	&	Unique Data	&	Resolution	&	Duration (s)	&	Distortion Type		&	Subjective Experiment	&	Annotators	\\
\midrule

KoNViD-1k \cite{hosu2017konstanz}	&	2017	&	VQA	&	Generic	&	1,200	&	1,200	&	540p	&	8	&	In-the-wild		&	Crowdsourced	&	642	\\
LIVE-VQC \cite{sinno2018large}	&	2018	&	VQA	&	Generic	&	585	&	585	&	240p-1080p	&	10	&	In-the-wild		&	Crowdsourced	&	4,776	\\
YouTube-UGC \cite{wang2019youtube}	&	2019	&	VQA	&	Generic	&	1,500	&	1,500	&	360p-4k	&	20	&	In-the-wild		&	Crowdsourced	&	-	\\
LSVQ \cite{ying2021patch}	&	2021	&	VQA	&	Generic	&	38,811	&	38,811	&	Diverse	&	5-12	&	In-the-wild		&	Crowdsourced	&	6,284	\\
FineVD \cite{duan2024finevq}	&	2025	&	VQA	&	Generic	&	6,104	&	6,104	&	Diverse	&	8	&	In-the-wild		&	In-lab	&	22	\\
\hdashline

GFIQA-20K \cite{su2023going}	&	2023	&	FIQA	&	Face	&	20,000	&	20,000	&	512×512	&	-	&	In-the-wild		&	In-lab	&	13	\\
PIQ23 \cite{chahine2023image}	&	2023	&	FIQA	&	Face	&	5,116	&	5,116	&	Diverse	&	-	&	In-capture		&	In-lab	&	30	\\
GFIQA-40K \cite{chen2024dsl}	&	2024	&	FIQA	&	Face	&	39,312	&	39,312	&	512×512	&	-	&	In-the-wild		&	-	&	20	\\
FIQA \cite{liu2024assessing}	&	2024	&	FIQA	&	Face	&	42,125	&	625	&	Diverse	&	-	&	Compression		&	Crowdsourced	&	1,432	\\
\hdashline
CFVQA \cite{li2024perceptual}	&	2024	&	FVQA	&	Face	&	3,240	&	135	&	512×512	&	5	&	Compression		&	In-lab	&	32	\\
\rowcolor[gray]{.92}
\textbf{FVQ-20K (Ours)}	&	2025	&	FVQA	&	Face	&	20,000	&	20,000	&	Diverse	&	5	&	In-the-wild		&	In-lab	&	35	\\

\bottomrule
\end{tabular}
\label{tab:dataset}
}
\vspace{-2mm}
\centering
\end{table*}

\begin{figure}
\centering
\includegraphics[width=\linewidth]{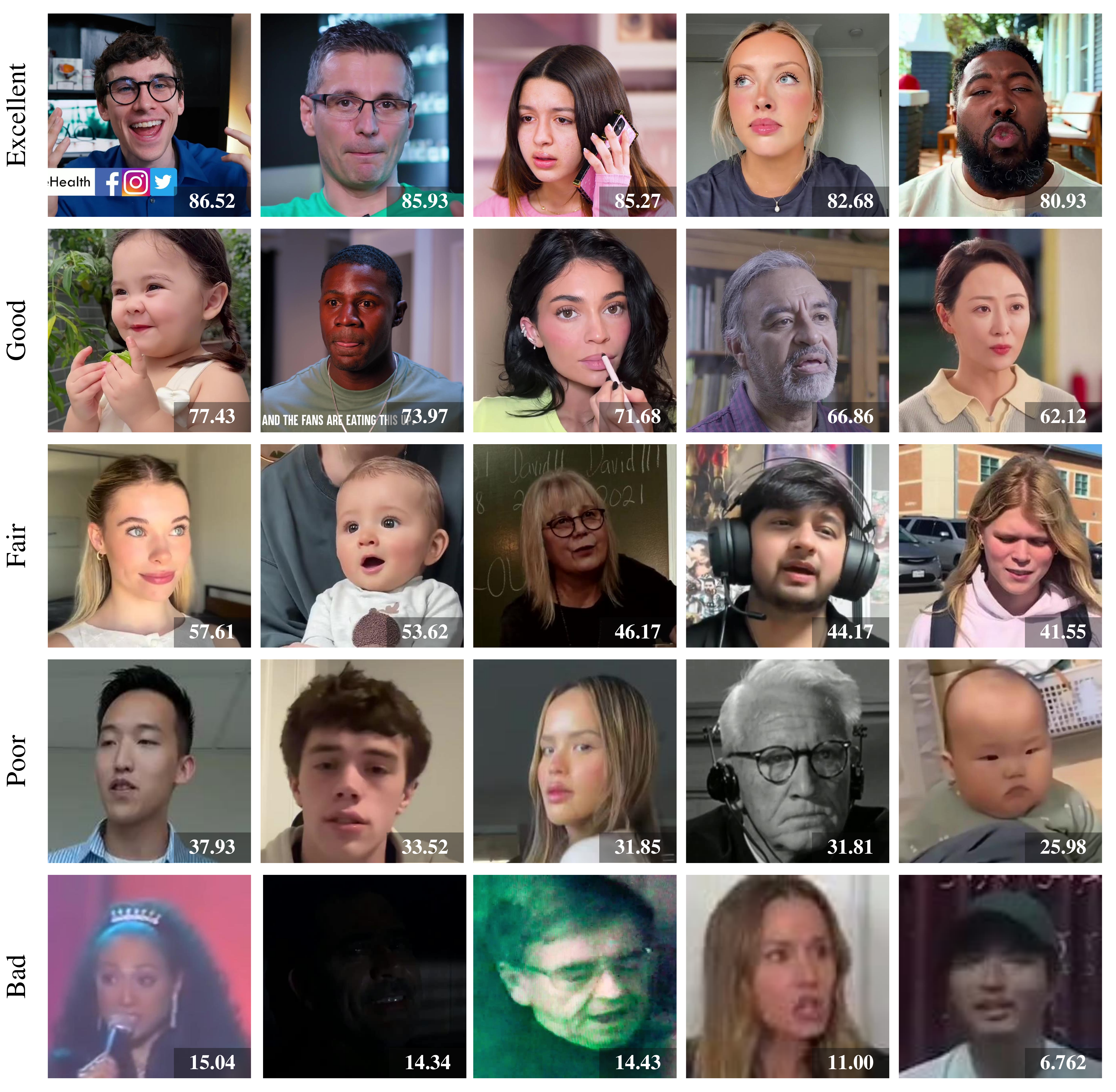}
\vspace{-6mm}
\caption{Sample video frames and corresponding MOSs from five quality levels of the proposed FVQ-20K dataset. We encourage readers to zoom-in for details.}
\vspace{-2mm}
\label{fig:demo}
\end{figure}

\begin{figure}
\centering
\includegraphics[width=\linewidth]{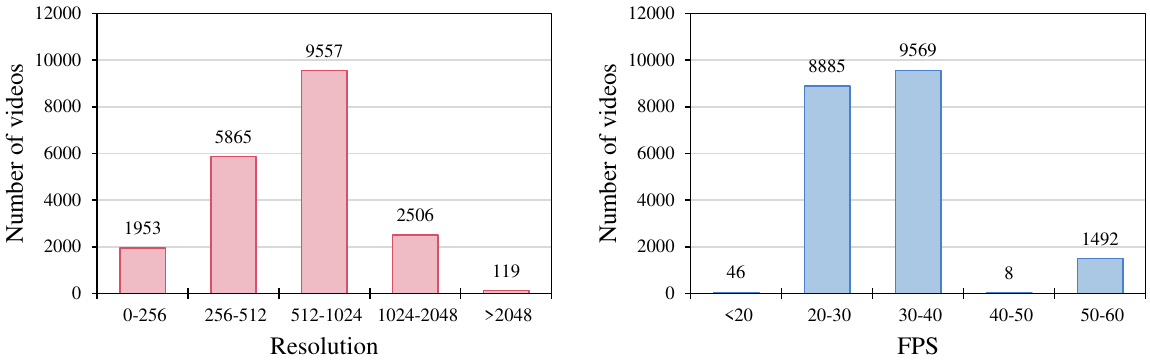}
\vspace{-6mm}
\caption{The bar charts of the resolution distribution (Left) and FPS distribution (Right) of the face videos in FVQ-20K.}
\vspace{-3mm}
\label{fig:video}
\end{figure}

\section{Introduction}
With the rapid development of network and video processing technologies, recent years have witnessed an explosion of social media applications with countless video content. 
However, these videos may suffer from various degradations caused by diverse acquisition and transmission conditions, thus video quality assessment (VQA) is becoming an increasingly popular topic in computer vision and multimedia with numerous research works \cite{ying2021patch,lu2024kvq,duan2024finevq}.
Compared to general video content, face and portrait videos play an even more significant role in the quality of experience (QoE), as human faces naturally attract more visual attention than backgrounds or non-human objects, making their visual quality more perceptually significant.
Analyzing and measuring the quality of face videos can help monitor and optimize user experience at social media sites \cite{ying2021patch,duan2024finevq}, promote the development of face-processing algorithms such as face video restoration, super-resolution, and enhancement, and ensure high-quality training data for generative models \cite{su2023going,chen2024dsl}.
However, face video quality assessment (FVQA) is rarely explored due to the absence of specific large-scale FVQA datasets.

Years of research on VQA have demonstrated promising results \cite{you2021long,chen2021learning,liao2022exploring,wu2022fast,sun2022deep,wu2023exploring,wu2023towards,yuan2023capturing,zhang2023spatial,sun2024analysis} thanks to the availability of various VQA datasets \cite{hosu2017konstanz,sinno2018large,wang2019youtube,ying2021patch,lu2024kvq,duan2024finevq}. 
However, these VQA datasets are designed for general videos, making the models devised on them less effective for face videos due to two aspects: (1) human perception of face video quality may differ from general video, as the HVS has specialized mechanisms for face processing \cite{kanwisher1997fusiform,haxby2000distributed,su2023going}, which is overlooked in general VQA methods; (2) face videos only account for a small fraction of general VQA datasets, which limits the generalization ability of VQA methods to face videos.
In recent years, some works have explored the problem of face image quality assessment from the aspects of either face recognition \cite{best2018learning,hernandez2019faceqnet,terhorst2020ser,ou2021sdd,boutros2023cr} or perceptual quality \cite{su2023going,chen2024dsl,chahine2023image,liu2024assessing}.
However, they focus solely on face images without considering temporal distortions, making them less effective when directly applied to face videos.
Recently, CFVQA \cite{li2024perceptual} has made the first attempt to explore the compressed FVQA problem. However, the dataset only contains 3,240 videos manually compressed from only 135 source videos using six video compression algorithms, which is not only limited in quantity and diversity but also fails to represent complex degradations in real-world scenarios. These limitations severely hinder the development of general FVQA algorithms.

To alleviate the absence of general FVQA datasets, we present \textbf{FVQ-20K}, the first large-scale in-the-wild FVQA dataset (see Figure \ref{fig:teaser} and Table \ref{tab:dataset}). Concretely, we collect face videos from both short-form video platform TikTok \cite{tiktok} and traditional video platform YouTube \cite{youtube} as shown in Figure \ref{fig:teaser} (a). For TikTok videos, we first collect over 100,000 videos across 11 categories related to people, and then filter and crop them to obtain 9,930 face videos, each with a fixed duration of 5 seconds. For YouTube videos, we integrate two public face video datasets, \textit{i.e.}, CelebV-HQ \cite{zhu2022celebv} and CelebV-Text \cite{yu2023celebv}, and obtain a total of 10,070 face videos by filtering out duplicate identities and videos shorter than 5 seconds. After these steps, we obtain a rich in-the-wild face video dataset with various degradations such as blur, shaking, \textit{etc.}, as shown in Figure \ref{fig:teaser} (b), and diverse face attributes \cite{serengil2024lightface,narayan2024facexformer} including gender, age, race, and emotion as shown in Figure \ref{fig:teaser} (c).
Furthermore, we recruit 35 subjects from the university to rate the videos, and a professional team of image processing researchers is responsible for volunteer training and monitoring the quality of the scores. 
As shown in Figure \ref{fig:teaser} (d), the FVQ-20K dataset is equipped with high-quality mean opinion scores (MOSs) to represent the perceptual quality of each face video.

Along with the FVQ-20K dataset, we propose a novel FVQA method named \textbf{FVQ-Rater} to achieve human-like rating and scoring of face videos, which enhances large multimodal models (LMMs) \cite{chen2024expanding} with elaborate multi-dimensional features and incorporates low-rank adaptation (LoRA) \cite{hu2022lora} technique for efficient instruction tuning \cite{liu2023visual,liu2024improved}.
Concretely, we utilize InternViT \cite{chen2024expanding} and SlowFast \cite{feichtenhofer2019slowfast} to extract general spatial and temporal features of face videos from key frames and entire video, respectively. Moreover, we also extract two face-specific features: (1) portrait features, which are extracted using InternViT \cite{chen2024expanding} only from the portrait part of face videos obtained by the background matting \cite{lin2022robust} technique, and (2) face embeddings, which encode high-level facial features such as geometric structure, appearance, relative positions of facial landmarks, \textit{etc.} \cite{schroff2015facenet}.
All these features, along with text embeddings, are integrated into a large language model (LLM) backbone for quality-oriented response and rating/scoring.
FVQ-Rater is trained through two stages: quality aware pre-training and MOS-oriented fine-tuning.
In the first stage, we construct 16,000 question-answer pairs related to the quality level of face videos, and train feature projectors to align face quality-related features with the pre-trained LLM \cite{chen2024expanding}.
In the second stage, we further fine-tune the vision encoder (\textit{i.e.}, InternViT), LLM backbone, and feature projectors to achieve accurate MOS regression through the quality regression module. The ViT and LLM are fine-tuned by incorporating LoRA weights to their pre-trained weights.
Extensive experiments demonstrate the superiority of our specialized FVQA method, FVQ-Rater.

In summary, our main contributions are:
\begin{itemize}
    \item We establish the first large-scale FVQA dataset, \textbf{FVQ-20K}, which contains 20,000 diverse face videos with MOS annotations.
    \item Based on the FVQ-20K dataset, we investigate and benchmark the performance of general-purpose VQA methods, FIQA methods, and state-of-the-art LMM models on the FVQA task.
    \item We propose \textbf{FVQ-Rater}, a novel LMM-based FVQA method that integrates spatial, temporal, and face-specific features, and adopts the LoRA-based instruction tuning technique, which is the first attempt to explore the potential of LMMs for the FVQA task.
    \item Extensive experiments on both FVQ-20K and CFVQA datasets demonstrate the superiority of the proposed FVQ-Rater method.
\end{itemize}

\begin{figure*}
\centering
\includegraphics[width=\linewidth]{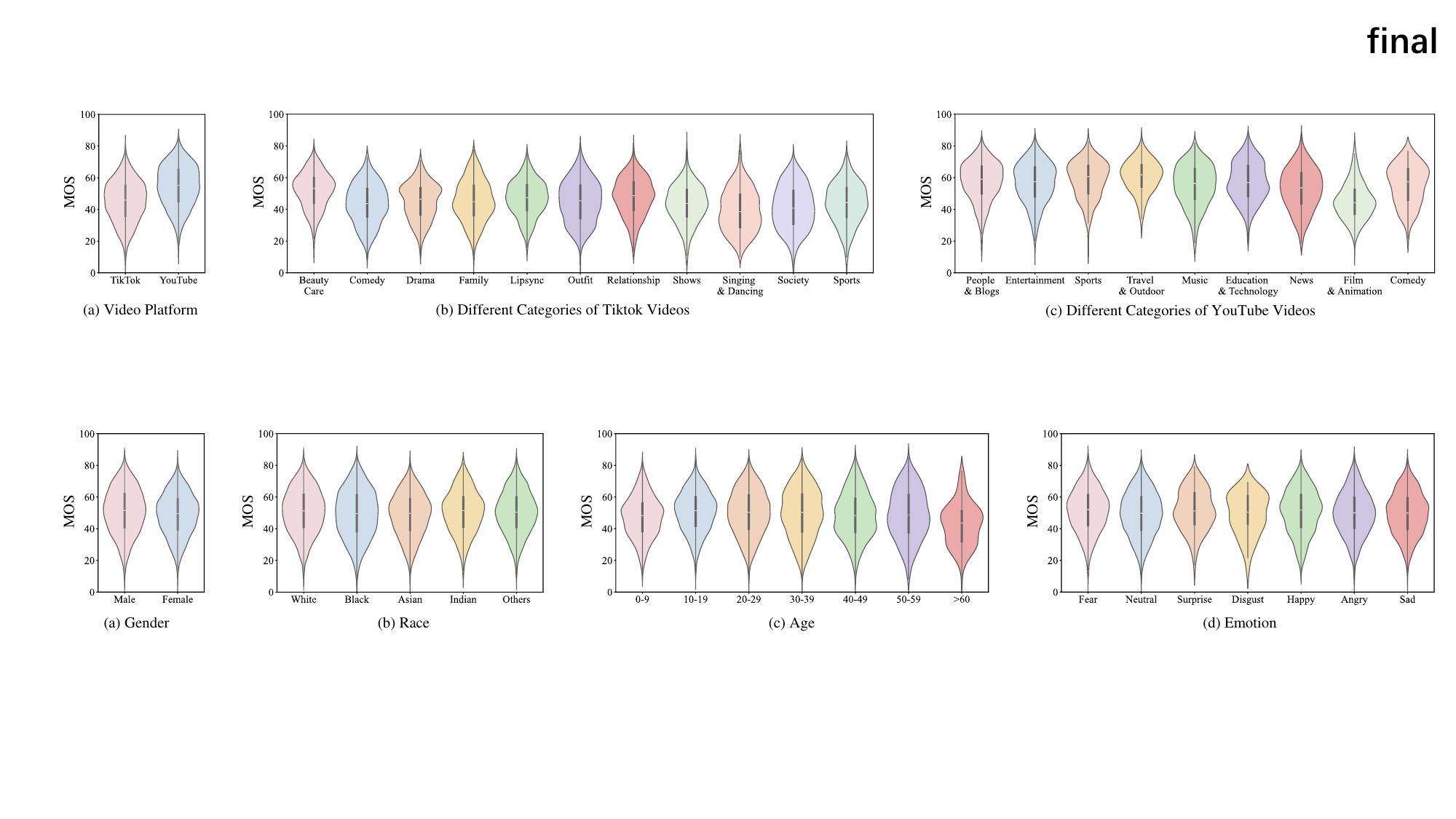}
\vspace{-6mm}
\caption{The MOS distributions in terms of different video source categories: (a) two video platforms, (b) eleven categories of TikTok videos, and (c) nine categories of YouTube videos.}
\vspace{-2mm}
\label{fig:platform_mos}
\end{figure*}

\begin{figure*}
\centering
\includegraphics[width=\linewidth]{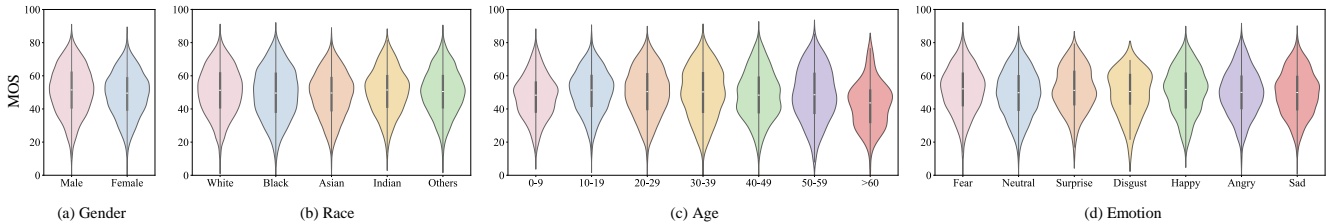}
\vspace{-7mm}
\caption{The MOS distributions in terms of different face attributes: (a) gender, (b) race, (c) age, and (d) emotion.}
\vspace{-3mm}
\label{fig:attribute_mos}
\end{figure*}

\section{Related Work}
\noindent\textbf{Face Quality Assessment.}
Face quality assessment contains face image quality assessment (FIQA) and face video quality assessment (FVQA).
FIQA is an increasingly popular topic which can be divided into two research areas \cite{chen2024dsl,su2023going,wang2022survey}, including biometric face image quality assessment (BFIQA) \cite{abdel2007application,gao2007standardization,abaza2012quality,best2018learning,zhao2019face,lijun2019multi,hernandez2019faceqnet,terhorst2020ser,meng2021magface,ou2021sdd,boutros2023cr} and generic face image quality assessment (GFIQA) \cite{su2023going,chen2024dsl,chahine2023image,liu2024assessing,jo2023ifqa}.
BFIQA aims at ensuring the quality of face images for the face recognition systems \cite{schlett2022face,ou2024refining}. Earlier BFIQA methods \cite{abdel2007application,gao2007standardization,abaza2012quality,lijun2019multi} usually evaluate the face image quality through hand-crafted features such as illumination, pose, facial expression, blur, occlusion, \textit{etc.} Recent deep learning-based methods measure the quality of face images in a data-driven manner, utilizing either pre-trained face recognition backbones \cite{terhorst2020ser,meng2021magface,boutros2023cr} or human annotations \cite{best2018learning,zhao2019face}. However, these methods do not focus on the perceptual quality of face images.
In contrast, GFIQA methods assess face images from the perspective of human perceptual quality, which is an emerging task \cite{su2023going}. Recently, impressive progress has been made in GFIQA thanks to the availability of several face image quality assessment datasets and advances in deep learning-based methods \cite{su2023going,chen2024dsl,chahine2023image,liu2024assessing}.
%% FVQA
Despite the extensive progress in FIQA, FVQA remains largely underexplored.
The only work focusing on FVQA explores the perceptual quality in face video compression scenarios \cite{li2024perceptual} by collecting a compressed face video assessment dataset termed CFVQA. However, CFVQA only contains 3,240 videos manually compressed from 135 source videos using six video compression algorithms, which has limited quantity, diversity, and degradation types. Moreover, the method FAVOR devised based on CFVQA \cite{li2024perceptual} is a full-reference (FR) VQA method, which is hard to be used in real-world applications.
To fill this gap, we construct the first large-scale in-the-wild FVQA dataset, FVQ-20K, and propose the first no-reference (NR) FVQA method, FVQ-Rater.

\noindent\textbf{Video Quality Assessment.}
Video quality assessment (VQA) is a long-standing problem with extensive research efforts and a variety of publicly available datasets.
VQA datasets can be categorized into three types according to the source of distortions in the videos, which includes captured VQA datasets \cite{nuutinen2016cvd2014,ghadiyaram2017capture}, UGC VQA datasets (in-the-wild VQA datasets) \cite{hosu2017konstanz,sinno2018large,wang2019youtube,ying2021patch,duan2024finevq}, and UGC+compression datasets \cite{li2020ugc,wang2021rich,yu2021predicting,zhang2023md,lu2024kvq}. However, all these datasets are designed for general-purpose videos and include very few face videos.
VQA methods can be divided into two categories, including traditional methods \cite{saad2014blind,mittal2015completely,korhonen2019two,tu2021rapique,kancharla2021completely,tu2021ugc,zheng2022completely} and learning-based methods \cite{you2021long,chen2021learning,liao2022exploring,wu2022fast,sun2022deep,wu2023exploring,wu2023towards,yuan2023capturing,sun2024analysis,xing2024clipvqa,ge2025lmm}. Traditional methods generally use hand-crafted features and predict quality scores through regression algorithms such as support vector regression (SVR) \cite{smola2004tutorial}, which struggle to handle complex real-world distortions. In recent years, learning-based methods have achieved superior performance through a data-driven manner. However, all these methods focus on general videos and do not consider human-specific features.
In contrast, the proposed FVQ-20K dataset and the FVQ-Rater method are both specialized for face videos.

\section{FVQ-20K Dataset}
To address the scarcity of datasets and advance the progress of FVQA, we present \textbf{FVQ-20K}, the first large-scale in-the-wild FVQA dataset as shown in Table \ref{tab:dataset}, which contains 20,000 diverse face videos with high-quality MOS annotations.
In this section, we introduce the dataset construction pipeline including face video collection (Section \ref{sec:video_collection}), subjective study (Section \ref{sec:subjective_study}), and subjective data processing (Section \ref{sec:subjective_data_processing}). The comprehensive analysis of the FVQ-20K dataset is detailed in Section \ref{sec:data_analysis}.

\subsection{Face Video Collection}
\label{sec:video_collection}
The main principle of the video collection process is to gather a wide range of face videos that encompass diverse and comprehensive in-the-wild distortions. To this end, we collect face videos from two popular social media platforms, \textit{i.e.}, a short-form video platform TikTok \cite{tiktok} and a traditional video platform YouTube \cite{youtube}.

For the TikTok \cite{tiktok} part, we select 11 categories of video content where people appear most frequently, including relationship, beauty care, family, lipsync, society, shows, singing \& dancing, outfit, sports, comedy, and drama, and download a total of 100,212 videos from these categories.
Then, we filter out videos that do not contain faces and crop the face regions from the remaining videos to 5-second face videos through the commonly used face and landmark detection models \cite{zhang2016joint,bulat2017far}. 
As a result, we obtain a total of 9,930 face videos from TikTok.

As for the YouTube \cite{youtube} part, since there are many publicly available portrait video datasets for other face-related tasks, we directly integrate and filter two recent datasets to construct the face video portion of our dataset.
Specifically, we merge the CelebV-HQ \cite{zhu2022celebv} dataset and the CelebV-Text \cite{yu2023celebv} dataset, and obtain a total of 104,218 face videos. Then we eliminate duplicate identities and uniform videos to 5 seconds to construct our final dataset.
Eventually, we obtain a total of 10,070 face videos from YouTube, which also span a variety of categories, as shown in Figure \ref{fig:teaser} (a).

% \subsection{Human Annotation}
\subsection{Subjective Study}
\label{sec:subjective_study}
To ensure the quality of subjective study results, we recruit 35 college students as subjects instead of crowdsourcing, and conduct the experiment following the recommendation of ITU-BT.500 \cite{series2012methodology}.

Considering the large scale of our dataset, we randomly divide the 20,000 videos into 40 batches to carry out the experiment.
For each video, subjects are asked to rate the perceptual quality on a scale of 0 to 5, where the intervals 0–1, 1–2, 2–3, 3–4, and 4–5 correspond to bad, poor, fair, good, and excellent, respectively.
Before the formal experiments, we train all the subjects by explaining the detailed rating criteria and numerous examples from each quality level. 
After the training session, we conduct a testing session with 15 face videos to evaluate whether the subjects are sufficiently trained. Only the subjects who pass the testing session are allowed to take part in the formal rating experiments. During the rating process, after completing each batch of data, we perform the subject rejection procedure \cite{series2012methodology} on each batch and exclude any subject identified as an outlier from participating in subsequent experiments.
Please refer to the \textit{Sup. Mat.} for more details.

\begin{figure*}[h]
\centering
\includegraphics[width=\linewidth]{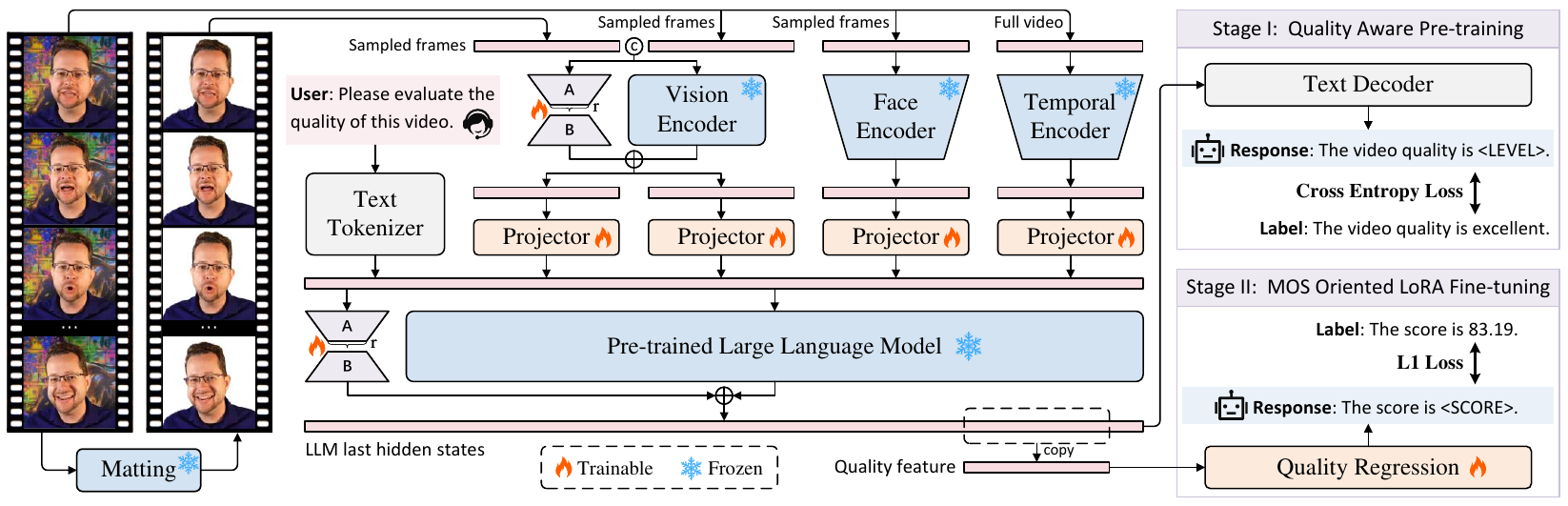}
\vspace{-7mm}
\caption{Overview of our FVQ-Rater method. Given face video and text prompt, FVQ-Rater first extracts multi-dimensional features through the vision encoder, face encoder, and temporal encoder, and then projects the features into the LLM input space through four different projectors. The feature embeddings are then combined with text embeddings and fed into the pre-trained LLM for further processing. The output features of the LLM are decoded to text output or partially fed into the quality regression module to predict the quality score. FVQ-Rater is trained in two stages: quality-aware pre-training via quality levels described in textual form and MOS-oriented LoRA fine-tuning using quality scores.}
\vspace{-3mm}
\label{fig:method}
\end{figure*}

\subsection{Subjective Data Processing}
\label{sec:subjective_data_processing}

We first perform outlier detection and subject rejection on all the 20,000 videos based on the guidelines provided by ITU-BT.500 \cite{series2012methodology}. As a result, no subjects are rejected, and about 3.05\% of the subjective scores are identified as outliers and are subsequently removed.
The remaining valid subjective scores are then converted into Z-scores by:
\begin{equation}
  z_{ij}=\frac{r_{ij}-\mu_{i}}{\sigma_{i}},
\end{equation}
where $r_{ij}$ is the raw score rated by the $i$-th subject to the $j$-th face video, $\mu_{i}$ and $\sigma_{i}$ represent the mean score and standard deviation given by subject $i$, respectively.
Subsequently, Z-scores are linearly scaled to the range of $[0, 100]$, assuming that the Z-scores of a subject follow a standard Gaussian distribution \cite{seshadrinathan2010study,li2024perceptual} within the range of $[-3, +3]$:
\begin{equation}
  z_{ij}'=\frac{100(z_{ij}+3)}{6}.
\end{equation}
Finally, the rescaled Z-scores $ z_{ij}'$ are averaged over subjects to obtain the mean opinion scores (MOSs):
\begin{equation}
  \text{MOS}_{j}=\frac{1}{N}\sum_{i=1}^{N}z_{ij}',
\end{equation}
where $j$ represents the $j$-th video, and $N$ is the number of subjects.

\subsection{Data Analysis}
\label{sec:data_analysis}
In this section, we provide comprehensive analyses of the proposed FVQ-20K dataset in terms of both face videos and their corresponding scores.
First of all, some demo frames and their MOSs are shown in Figure \ref{fig:demo}. Please also refer to the \textit{Sup. Mat.} for more demos. The MOS distribution of the entire FVQ-20K dataset is shown in Figure \ref{fig:teaser} (d), which spans a wide range.
It is obvious that videos with more severe distortions tend to receive lower scores.
The face videos in the FVQ-20K dataset are as 5-second square videos, while the resolution, encoding format, and frames per second (FPS) remain unchanged to ensure the save quality as the original in-the-wild videos. The resolution and FPS distributions of all the face videos are shown in Figure \ref{fig:video}, which exhibit great diversity.

Beyond resolution and FPS, the content of the face videos is also highly diverse, not only in terms of the platforms they originate from but also in terms of the facial attributes captured in the videos.
As shown in Figure \ref{fig:teaser} (a) and (c), the face videos in the FVQ-20K dataset come from up to $11+9=20$ categories on TikTok \cite{tiktok} and YouTube \cite{youtube}, and the faces in the videos show a wide diversity in gender, race, age, and emotion.
To this end, we further analyze the MOS distribution under each category based on the aforementioned classification criteria. The MOS distributions for two online video platforms, eleven TikTok video categories, and nine YouTube categories are shown in Figure \ref{fig:platform_mos}. It can be observed that the overall quality of face videos from YouTube is higher than those from TikTok. The MOS distributions vary across different video content categories, particularly the beauty care content in TikTok videos and film \& animation content in YouTube videos, which also demonstrates the diversity of our dataset.
Similarly, we visualize the MOS distributions for different genders, races, ages, and emotions in Figure \ref{fig:attribute_mos}. It can be observed that the MOS distributions across different genders, races, and emotions are generally similar, indicating that our MOSs are well-focused on face video quality and are not affected by other subjective preferences. When it comes to different age groups, the MOS distribution of face videos featuring individuals over the age of 60 differs from that of other age groups, and the overall quality is also relatively lower.

\begin{table*}[h]
\centering
\renewcommand\arraystretch{0.9}
\caption{Quality score prediction performance of state-of-the-art methods on the proposed FVQ-20K dataset (including its TikTok and YouTube subsets) and the CFVQA \cite{li2024perceptual} dataset. $\heartsuit$, $\spadesuit$, \raisebox{0.4ex}{\scalebox{0.6}{$\bigcirc$}}, \scalebox{0.8}{\ding{170}}, $\diamondsuit$, and $\clubsuit$ denote traditional VQA methods, learning-based VQA methods, FIQA methods, FBP methods, LMM-based methods, and FVQA method, respectively. * and \dag indicate models trained and fine-tuned on the corresponding datasets, respectively. Note that FAVOR \cite{li2024perceptual} is a full-reference method and is not applicable to the in-the-wild FVQ-20K dataset. The best and runner-up performances are bold and underlined, respectively.}
\vspace{-2mm}

\resizebox{\linewidth}{!}{
\begin{tabular}{lcccccccccccc}
\toprule
Dataset  & \multicolumn{3}{c}{Our FVQ-20K (TikTok)}  & \multicolumn{3}{c}{Our FVQ-20K (YouTube)}  & \multicolumn{3}{c}{Our FVQ-20K}  & \multicolumn{3}{c}{CFVQA} \\ \cmidrule(lr){2-4} \cmidrule(lr){5-7} \cmidrule(lr){8-10} \cmidrule(lr){11-13}

Method / Metric  & SRCC\,$\uparrow$  & PLCC\,$\uparrow$  & KRCC\,$\uparrow$  & SRCC\,$\uparrow$  & PLCC\,$\uparrow$  & KRCC\,$\uparrow$  & SRCC\,$\uparrow$  & PLCC\,$\uparrow$  & KRCC\,$\uparrow$  & SRCC\,$\uparrow$  & PLCC\,$\uparrow$  & KRCC\,$\uparrow$\\ 
\midrule

$\heartsuit$ RAPIQUE \cite{tu2021rapique}	&	0.6657 	&	0.7126 	&	0.4842 	&	0.6442 	&	0.6709 	&	0.4577 	&	0.6720 	&	0.7065 	&	0.4865 	&	0.5636 	&	0.5600 	&	0.3876 	\\
$\heartsuit$ VIDEVAL \cite{tu2021ugc}	&	0.8310 	&	0.8422 	&	0.6456 	&	0.7275 	&	0.7180 	&	0.5352 	&	0.7994 	&	0.8028 	&	0.6124 	&	0.5058 	&	0.5208 	&	0.3473 	\\
$\spadesuit$ VSFA \cite{li2019quality}*	&	0.8992 	&	0.9032 	&	0.7245 	&	0.8878 	&	0.8896 	&	0.7069 	&	0.9040 	&	0.9080 	&	0.7319 	&	0.9166 	&	0.9092 	&	0.7391 	\\
$\spadesuit$ GSTVQA \cite{chen2021learning}*	&	0.9032 	&	0.9068 	&	0.7296 	&	0.8979 	&	0.9012 	&	0.7200 	&	0.8975 	&	0.9048 	&	0.7233 	&	0.9451 	&	0.9546 	&	0.7920 	\\
$\spadesuit$ SimpleVQA \cite{sun2022deep}*	&	0.8361 	&	0.8212 	&	0.6381 	&	0.8708 	&	0.8662 	&	0.6790 	&	0.8763 	&	0.8604 	&	0.6843 	&	0.7583 	&	0.7578 	&	0.5510 	\\
$\spadesuit$ FastVQA \cite{wu2022fast}*	&	0.9013 	&	0.8970 	&	0.7281 	&	0.8793 	&	0.8817 	&	0.6970 	&	0.8977 	&	0.8993 	&	0.7229 	&	0.9461 	&	0.9581 	&	0.7973 	\\
$\spadesuit$ Dover \cite{wu2023exploring}*	&	0.9145 	&	0.9114 	&	0.7430 	&	0.8928 	&	0.8909 	&	0.7100 	&	0.9124 	&	0.9097 	&	0.7407 	&	\underline{0.9568} 	&	\underline{0.9588} 	&	\underline{0.8157} 	\\
$\spadesuit$ KSVQE \cite{lu2024kvq}*	&	0.9230 	&	0.9159 	&	0.7589 	&	0.9063 	&	0.8991 	&	0.7307 	&	\underline{0.9244} 	&	0.9188 	&	0.7600 	&	0.9510 	&	0.9516 	&	0.8015 	\\

\hdashline
\raisebox{0.4ex}{\scalebox{0.6}{$\bigcirc$}} SDD-FIQA \cite{ou2021sdd}*	&	0.7578 	&	0.7678 	&	0.5619 	&	0.6805 	&	0.7093 	&	0.4941 	&	0.7537 	&	0.7696 	&	0.5604 	&	0.3214 	&	0.3247 	&	0.2258 	\\
\raisebox{0.4ex}{\scalebox{0.6}{$\bigcirc$}} DSL-FIQA \cite{chen2024dsl}*	&	\underline{0.9235} 	&	\underline{0.9250} 	&	\underline{0.7592} 	&	\underline{0.9079} 	&	\underline{0.9101} 	&	\underline{0.7350} 	&	0.9243 	&	\underline{0.9269} 	&	\underline{0.7620} 	&	0.5857 	&	0.5959 	&	0.4262 	\\
\scalebox{0.8}{\ding{170}} REX-INCEP \cite{bougourzi2022deep}*	&	0.9130 	&	0.9171 	&	0.7467 	&	0.9019 	&	0.9052 	&	0.7270 	&	0.9174 	&	0.9212 	&	0.7528 	&	0.4154 	&	0.4333 	&	0.3143 	\\
\scalebox{0.8}{\ding{170}} 2D\_FAP \cite{liu2025lightweight}*	&	0.6448 	&	0.6655 	&	0.4591 	&	0.6783 	&	0.6786 	&	0.4887 	&	0.6653 	&	0.6734 	&	0.4765 	&	0.4229 	&	0.4303 	&	0.3012 	\\
\scalebox{0.8}{\ding{170}} MEBeauty \cite{lebedeva2022mebeauty}*	&	0.7945 	&	0.7976 	&	0.5976 	&	0.7971 	&	0.8053 	&	0.6052 	&	0.7954 	&	0.8023 	&	0.6015 	&	0.4278 	&	0.4299 	&	0.3064 	\\

\hdashline
$\diamondsuit$ Video-LLaVA-7B \cite{lin2023video}	&	0.1338 	&	0.1463 	&	0.0989 	&	0.0667 	&	0.0950 	&	0.0494 	&	0.0766 	&	0.0968 	&	0.0565 	&	0.0783 	&	0.0613 	&	0.0583 	\\
$\diamondsuit$ VideoLLaMA2-7B \cite{cheng2024videollama}	&	0.0265 	&	0.0332 	&	0.0206 	&	0.0974 	&	0.0136 	&	0.0764 	&	0.0936 	&	0.0358 	&	0.0731 	&	0.1408 	&	0.1459 	&	0.1107 	\\
$\diamondsuit$ VideoLLaMA3-7B \cite{zhang2025videollama}	&	0.0762 	&	0.0209 	&	0.0548 	&	0.1618 	&	0.0043 	&	0.1180 	&	0.1158 	&	0.0047 	&	0.0838 	&	0.1356 	&	0.0373 	&	0.0982 	\\
$\diamondsuit$ Qwen2-VL-7B \cite{wang2024qwen2}	&	0.1679 	&	0.0013 	&	0.1266 	&	0.1389 	&	0.1339 	&	0.1042 	&	0.0073 	&	0.0190 	&	0.0029 	&	0.0128 	&	0.0460 	&	0.0078 	\\
$\diamondsuit$ Qwen2-VL-7B-Instruct \cite{wang2024qwen2}	&	0.4262 	&	0.2631 	&	0.3282 	&	0.2037 	&	0.2214 	&	0.1526 	&	0.3783 	&	0.2705 	&	0.2875 	&	0.3259 	&	0.3935 	&	0.2508 	\\
$\diamondsuit$ Qwen2.5-VL-7B-Instruct \cite{bai2025qwen2}	&	0.3519 	&	0.2915 	&	0.2710 	&	0.2092 	&	0.1155 	&	0.1664 	&	0.3071 	&	0.2219 	&	0.2398 	&	0.0271 	&	0.0142 	&	0.0227 	\\
$\diamondsuit$ Deepseek-VL-7B-Chat \cite{lu2024deepseek}	&	0.0156 	&	0.0440 	&	0.0115 	&	0.1830 	&	0.0877 	&	0.1322 	&	0.1304 	&	0.0819 	&	0.0938 	&	0.0887 	&	0.0251 	&	0.0657 	\\
$\diamondsuit$ InternVL2.5-8B \cite{chen2024expanding}	&	0.0777 	&	0.1446 	&	0.0401 	&	0.0120 	&	0.0622 	&	0.0163 	&	0.0589 	&	0.1410 	&	0.0284 	&	0.0149 	&	0.1835 	&	0.0080 	\\
$\diamondsuit$ InternVL2.5-8B-MPO \cite{wang2024enhancing}	&	0.4304 	&	0.3637 	&	0.3414 	&	0.1008 	&	0.0479 	&	0.0617 	&	0.1040 	&	0.0493 	&	0.0988 	&	0.2831 	&	0.1858 	&	0.2165 	\\
$\diamondsuit$ Q-Align \cite{wu2023q}	&	0.8712 	&	0.8686 	&	0.6916 	&	0.8582 	&	0.8609 	&	0.6726 	&	0.8715 	&	0.8716 	&	0.6921 	&	0.6890 	&	0.6522 	&	0.4962 	\\
$\diamondsuit$ Qwen2-VL-7B-Instruct \cite{wang2024qwen2}\dag	&	0.8923 	&	0.8866 	&	0.7235 	&	0.8856 	&	0.8849 	&	0.7127 	&	0.9015 	&	0.8984 	&	0.7353 	&	0.9367 	&	0.9363 	&	0.7795 	\\
$\diamondsuit$ Qwen2.5-VL-7B-Instruct \cite{bai2025qwen2}\dag	&	0.9042 	&	0.9045 	&	0.7429 	&	0.8939 	&	0.8957 	&	0.7242 	&	0.9122 	&	0.9115 	&	0.7525 	&	0.9274 	&	0.9302 	&	0.7632 	\\
$\diamondsuit$ InternVL2.5-8B \cite{chen2024expanding}\dag	&	0.9019 	&	0.8936 	&	0.7404 	&	0.8944 	&	0.8825 	&	0.7291 	&	0.9090 	&	0.9011 	&	0.7511 	&	0.8933 	&	0.8903 	&	0.7178 	\\
$\diamondsuit$ InternVL2.5-8B-MPO \cite{wang2024enhancing}\dag	&	0.8789 	&	0.8293 	&	0.7212 	&	0.8175 	&	0.6628 	&	0.6695 	&	0.8545 	&	0.7634 	&	0.7095 	&	0.8510 	&	0.8047 	&	0.6725 	\\

\hdashline
$\clubsuit$ FAVOR \cite{li2024perceptual}*	&	{\small N/A}	&	{\small N/A}	&	{\small N/A}	&	{\small N/A}	&	{\small N/A}	&	{\small N/A}	&	{\small N/A}	&	{\small N/A}	&	{\small N/A}	&	0.9060 	&	0.9229 	&	0.7248 	\\
\rowcolor[gray]{.92}
\textbf{FVQ-Rater (Ours)}*	&	\textbf{0.9326} 	&	\textbf{0.9334} 	&	\textbf{0.7786} 	&	\textbf{0.9297} 	&	\textbf{0.9279} 	&	\textbf{0.7717} 	&	\textbf{0.9388} 	&	\textbf{0.9382} 	&	\textbf{0.7887} 	&	\textbf{0.9679} 	&	\textbf{0.9717} 	&	\textbf{0.8423} 	\\

\bottomrule
\end{tabular}
}
\vspace{-1mm}
\label{tab:compare_mos}
\end{table*}
% \vspace{-12pt}

\section{FVQ-Rater Method}
Along with the FVQ-20K dataset, we further propose \textbf{FVQ-Rater}, a novel LMM-based method specifically designed for the FVQA task.
As shown in Figure \ref{fig:method}, FVQ-Rater takes face video and user prompt as input, and outputs the quality level and quality score of the face video. The architecture and training strategy of FVQ-Rater are detailed in Section \ref{sec:architecture} and Section \ref{sec:training}, respectively.

\subsection{Architecture}
\label{sec:architecture}
\noindent\textbf{Spatial Feature Extraction.}
The spatial features are extracted from the key frames of the input face video through a vision encoder.
Specifically, we sample one frame per second from the input face video $\boldsymbol{V}$ based on its FPS, resulting in a total of 5 key frames $V_k$ for each video.
Then, we utilize a pre-trained vision transformer (ViT), \textit{i.e.}, InternViT \cite{chen2024expanding}, as the vision encoder $\mathcal{E}_I$.
To align the extracted spatial features with the input space of the pre-trained large language model (LLM), we adopt a 2-layer MLP projector $\mathcal{P}_S$ to map the spatial features into the textual embedding space.
The full process can be formulated as:
\begin{equation}
    F_s = \mathcal{P}_S(\mathcal{E}_I(V_k)),
\end{equation}
where $F_s$ is the projected spatial features compatible with the LLM input space.

\noindent\textbf{Temporal Feature Extraction.}
Considering that the spatial features extracted from sparse frames fail to capture complete temporal information, we extract temporal features from the continuous video frames.
Specifically, we utilize the pre-trained SlowFast network $\mathcal{E}_T$ as our temporal encoder to extract temporal features from the entire video $V$. A 2-layer MLP projector $\mathcal{P}_T$ is also adopted to project the temporal features to the LLM input space:
\begin{equation}
    F_t = \mathcal{P}_T(\mathcal{E}_T(V)),
\end{equation}
where $F_t$ is the projected temporal features compatible with the LLM input space.

\noindent\textbf{Face-Specific Feature Extraction.}
To better align with human perception of face videos, we further design two types of face-specific features inspired by the mechanisms of the human visual system: (1) portrait features extracted from the foreground portrait video, and (2) face embeddings that encode high-level facial features.

The portrait features are extracted from the key frames of the portrait video through the same vision encoder $\mathcal{E}_I$ employed for the spatial features. Specifically, the portrait video is obtained by separating the foreground from the background using a state-of-the-art video matting method \cite{lin2022robust}. Similar to the original video, we sample 5 key frames from the portrait video and feed them into the vision encoder to extract features only for people, which simulates the special sensitivity of the human visual system to people in the scene. A 2-layer MLP projector $\mathcal{P}_P$ is also employed to project the portrait features into the input embedding space of the LLM.
The complete process is formulated as:
\begin{equation}
    F_p = \mathcal{P}_P(\mathcal{E}_I(\mathcal{M}(V_k))),
\end{equation}
where $\mathcal{M}$ denotes the pre-trained matting network, and $F_p$ is the projected portrait features compatible with the LLM input space.

The face embeddings are extracted from the key frames of the original face video using a face encoder.
Specifically, we uniformly sample 10 key frames $V_f$ from the input video and then extract face embeddings from these frames using a pre-trained face recognition network, FaceNet \cite{schroff2015facenet}. The face embeddings encode high-level identity-related facial features such as geometric structure, appearance, relative positions of facial landmarks, which provide meaningful cues for the FVQA task, particularly in determining whether the facial structures encounter degradation and whether temporal distortions led to such degradation. Such face embeddings also simulate the human visual system’s ability to perceive, abstract, and distinguish facial identities, which helps achieve human-like quality perception.
Similarly, a 2-layer MLP projector $\mathcal{P}_F$ is also employed to align the face embeddings with the LLM input space:
\begin{equation}
    F_f = \mathcal{P}_F(\mathcal{E}_F(V_f)),
\end{equation}
where $\mathcal{E}_F$ refers to the pre-trained FaceNet \cite{schroff2015facenet}, and $F_f$ is the projected face embeddings compatible with the LLM input space.

\noindent\textbf{Feature Fusion via LLM.}
The aforementioned four types of features, combined with text embeddings, are fed into a pre-trained LLM for multimodal feature fusion and reasoning to support two downstream tasks: (1) quality-aware question answering, and (2) quality score prediction.
Concretely, user prompt and feature placeholder tokens are converted into textual embeddings through a text tokenizer and an embedding layer. The embeddings at the placeholder token position are then replaced with the extracted features $F_s$, $F_t$, $F_p$, and $F_f$.
Subsequently, we use InternLM2.5 \cite{chen2024expanding} as the pre-trained LMM, which takes the embeddings as input and outputs processed features for downstream tasks.
The LLM output features, \textit{i.e.}, the last hidden status of the LLM transformers, are decoded to the text output through the text decoder. 
Moreover, the quality-related part of the LLM output features is fed into the quality regression module to predict the MOS value. The quality regression module is built upon a 5-layer MLP with ReLU activation \cite{nair2010rectified} after each layer.

\subsection{Training}
\label{sec:training}
The FVQ-Rater is trained in two stages to achieve better performance on the quality-related tasks.

\noindent\textbf{Stage I: Quality-Aware Pre-training.}
To align multiple features extracted from the input face video with the pre-trained LLM, and let LMM understand the quality assessment task, we train the projectors $\mathcal{P}_S$, $\mathcal{P}_T$, $\mathcal{P}_P$, and $\mathcal{P}_F$ using (face video, question, answer) triplets and cross entropy loss, while keeping the vision encoder and LLM frozen. 
The question-answer pairs are in the form of: ``Question: \textit{Please evaluate the quality of this video.} Answer: \textit{The video quality is excellent.}"
The projectors $\mathcal{P}_S$ and $\mathcal{P}_P$ are initialized using the pre-trained MLP projector of InternVL2.5 \cite{chen2024expanding}, and other projectors are initialized randomly.
Through the quality-aware pre-training, the pre-trained LLM gains a better understanding of the multimodal input features and becomes more effective at extracting quality-related information from them.

\noindent\textbf{Stage II: MOS-Oriented LoRA Fine-tuning.}
To achieve an accurate quality score regression, we further train the quality regression module, and fine-tune the pre-trained projectors, vision encoder, and LLM.
Specifically, we use the L1 loss between the predicted quality score and the ground truth MOS to supervise the training process.
The regression module is randomly initialized, while other modules are pre-trained.
Besides, the vision encoder and LLM are fine-tuned using the LoRA technique \cite{hu2022lora}, which updates the pre-trained weight matrix $\boldsymbol{W} \in \mathbb{R}^{d\times k}$ by modeling the weight change with a low-rank decomposition $\Delta \boldsymbol{W}=\boldsymbol{B}\boldsymbol{A}$, where $\boldsymbol{B}\in \mathbb{R}^{d\times r}$ and $\boldsymbol{A}\in \mathbb{R}^{r\times k}$ are the trainable parameter matrices, and $r\ll\{d,k\}$ for efficiency. Then the model output $\boldsymbol{y}$ can be formulated as:
\begin{equation}
    \boldsymbol{y}=\boldsymbol{W}\boldsymbol{x}+\Delta \boldsymbol{W}\boldsymbol{x}=\boldsymbol{W}\boldsymbol{x}+\boldsymbol{B}\boldsymbol{A}\boldsymbol{x},
\end{equation}
where $\boldsymbol{x}$ denotes the model input.
Through the fine-tuning process, FVQ-Rater is able to predict quality scores for the input face videos.

\begin{table}[t]
\centering
\renewcommand\arraystretch{0.9}
\caption{Quality level classification performance of state-of-the-art LMMs and LMM-based VQA methods on the proposed FVQ-20K dataset (including its TikTok and YouTube subsets) and the CFVQA \cite{li2024perceptual} dataset. \dag indicate models fine-tuned on the corresponding datasets. The best and runner-up performances are bold and underlined, respectively.}
\vspace{-2.3mm}
\resizebox{\linewidth}{!}{
\begin{tabular}{lcccc}
\toprule

Method / Dataset  & \makecell[c]{FVQ-20K\\(TikTok)}  & \makecell[c]{FVQ-20K\\(YouTube)}  &FVQ-20K  & CFVQA \\ 

\midrule

Video-LLaVA-7B \cite{lin2023video}	&	3.290\%	&	6.290\%	&	4.800\%	&	3.290\%	\\
VideoLLaMA2-7B \cite{cheng2024videollama}	&	23.09\%	&	35.70\%	&	29.43\%	&	32.85\%	\\
VideoLLaMA3-7B \cite{zhang2025videollama}	&	25.03\%	&	43.25\%	&	34.20\%	&	28.54\%	\\
Qwen2-VL-7B \cite{wang2024qwen2}	&	14.77\%	&	21.19\%	&	18.00\%	&	11.91\%	\\
Qwen2-VL-7B-Instruct \cite{wang2024qwen2}	&	34.70\%	&	48.54\%	&	41.67\%	&	27.31\%	\\
Qwen2.5-VL-7B-Instruct \cite{bai2025qwen2}	&	15.77\%	&	38.08\%	&	27.00\%	&	24.23\%	\\
Deepseek-VL-7B-Chat \cite{lu2024deepseek}	&	43.22\%	&	36.62\%	&	39.90\%	&	34.50\%	\\
InternVL2.5-8B \cite{chen2024expanding}	&	45.84\%	&	52.32\%	&	49.10\%	&	30.80\%	\\
InternVL2.5-8B-MPO \cite{wang2024enhancing}	&	34.56\%	&	41.52\%	&	38.07\%	&	24.02\%	\\
Q-Align \cite{wu2023q}	&	14.30\%	&	20.26\%	&	17.30\%	&	9.450\%	\\

\hdashline
Qwen2-VL-7B-Instruct \cite{wang2024qwen2}\dag	&	76.31\%	&	74.70\%	&	75.50\%	&	\underline{81.72\%}	\\
Qwen2.5-VL-7B-Instruct \cite{bai2025qwen2}\dag	&	\underline{79.19\%}	&	78.61\%	&	78.90\%	&	79.88\%	\\
InternVL2.5-8B \cite{chen2024expanding}\dag	&	78.79\%	&	\underline{79.34\%}	&	\underline{79.07\%}	&	78.03\%	\\
InternVL2.5-8B-MPO \cite{wang2024enhancing}\dag	&	76.85\%	&	76.89\%	&	76.87\%	&	77.41\%	\\
\rowcolor[gray]{.92}
\textbf{FVQ-Rater (Ours)}	&	\textbf{80.94\%}	&	\textbf{81.06\%}	&	\textbf{81.00\%}	&	\textbf{86.86\%}	\\

\bottomrule
\end{tabular}
}
\vspace{-1.3mm}
\label{tab:compare_level}
\end{table}
\begin{table}[t]
\centering
\renewcommand\arraystretch{0.9}
\caption{Cross-dataset evaluation results. The models are trained on FVQ-20K and tested on CFVQA \cite{li2024perceptual}, or vice versa. The best performances are bold.}
\vspace{-2.3mm}
\resizebox{\linewidth}{!}{
\begin{tabular}{lcccccc}
\toprule
Dataset  & \multicolumn{3}{c}{FVQ-20K $\rightarrow$ CFVQA}  & \multicolumn{3}{c}{CFVQA $\rightarrow$ FVQ-20K} \\ \cmidrule(lr){2-4} \cmidrule(lr){5-7}

Method / Metric  & SRCC\,$\uparrow$  & PLCC\,$\uparrow$  & KRCC\,$\uparrow$  & SRCC\,$\uparrow$  & PLCC\,$\uparrow$  & KRCC\,$\uparrow$ \\ 
\midrule

SimpleVQA \cite{sun2022deep}	& 0.7374 	&	0.7527 	&	0.5353 	&	0.3309 	&	0.3498 	&	0.2275 	\\
FastVQA	\cite{wu2022fast} & 0.6359 	&	0.6263 	&	0.4610 	&	0.1041 	&	0.1028 	&	0.0706 	\\
Dover \cite{wu2023exploring}	&	0.6974 	&	0.6866 	&	0.5061 	&	0.7638 	&	0.7601 	&	0.5678 	\\
KSVQE \cite{lu2024kvq}	&	0.7742 	&	0.7736 	&	0.5714 	&	0.8096 	&	0.8098 	&	0.6145 	\\
DSL-FIQA \cite{chen2024dsl}	&	0.4096 	&	0.4406 	&	0.2881 	&	0.6372 	&	0.6599 	&	0.4644 	\\
\rowcolor[gray]{.92}
\textbf{FVQ-Rater (Ours)}	&	\textbf{0.8081} 	&	\textbf{0.7972} 	&	\textbf{0.6061} 	&	\textbf{0.8429} 	&	\textbf{0.8401} 	&	\textbf{0.6554} 	\\

\bottomrule
\end{tabular}
}
\vspace{-3mm}
\label{tab:cross}
\end{table}

\begin{table*}
%\vspace{-8mm}
\centering
\renewcommand\arraystretch{0.9}
\caption{Ablation study of FVQ-Rater on our FVQ-20K dataset and the CFVQA \cite{li2024perceptual} dataset.}
\vspace{-2mm}

% \vspace{-0.5em}
\resizebox{\textwidth}{!}{
\scriptsize
\begin{tabular}{ccccccc cccccc}
\toprule
\multicolumn{7}{c}{Feature \& Strategy}  & \multicolumn{3}{c}{FVQ-20K}  & \multicolumn{3}{c}{CFVQA} \\
\cmidrule(lr){1-7} \cmidrule(lr){8-10} \cmidrule(lr){11-13}
Spatial  & Portrait  & FaceNet  & Temporal  & LLM LoRA  & ViT LoRA  & Pretrain  & SRCC  & PLCC  & KRCC  & SRCC  & PLCC  & KRCC \\
\midrule

\checkmark	&		&		&		&		&		&		&	0.9132 	&	0.9144 	&	0.7449 	&	0.9528 	&	0.9501 	&	0.8086 	\\
\checkmark	&		&		&		&	\checkmark	&		&		&	0.9170 	&	0.9182 	&	0.7520 	&	0.9563 	&	0.9558 	&	0.8182 	\\
\checkmark	&		&		&		&	\checkmark	&	\checkmark	&		&	0.9283 	&	0.9241 	&	0.7707 	&	0.9650 	&	0.9648 	&	0.8387 	\\
\checkmark	&		&		&		&	\checkmark	&	\checkmark	&	\checkmark	&	0.9296 	&	0.9267 	&	0.7730 	&	0.9650 	&	0.9669 	&	0.8347 	\\
\checkmark	&	\checkmark	&		&		&	\checkmark	&	\checkmark	&	\checkmark	&	0.9367 	&	0.9345 	&	0.7849 	&	0.9666 	&	0.9674 	&	0.8390 	\\
\checkmark	&	\checkmark	&	\checkmark	&		&	\checkmark	&	\checkmark	&	\checkmark	&	0.9377 	&	0.9370 	&	0.7869 	&	0.9668 	&	0.9701 	&	0.8395 	\\
\rowcolor[gray]{.92}
\checkmark	&	\checkmark	&	\checkmark	&	\checkmark	&	\checkmark	&	\checkmark	&	\checkmark	&	\textbf{0.9388} 	&	\textbf{0.9382} 	&	\textbf{0.7887} 	&	\textbf{0.9679} 	&	\textbf{0.9717} 	&	\textbf{0.8423} 	\\

\bottomrule
\end{tabular}
\label{tab:ablation}
}
\vspace{-3mm}
\centering
\end{table*}

\section{Experiments}
\subsection{Implementation Details}
\noindent\textbf{Evaluation Datasets.}
To the best of our knowledge, there are only two FVQA datasets: CFVQA \cite{li2024perceptual} and our newly proposed FVQ-20K. We use both datasets in our experiments.
Both datasets are randomly split into training, validation, and test sets with a ratio of $80\%:5\%:15\%$. Concretely, CFVQA dataset contains 3,240 videos, which is divided into 2,592 for training, 161 for validation, and 487 for testing. FVQ-20K contains 20,000 videos, with 16,000 for training, 1,000 for validation, and 3,000 for testing.

\noindent\textbf{Training Details of FVQ-Rater.}
We use the pre-trained InternViT and InternLM2.5 from InternVL2.5-8B \cite{chen2024expanding} as our vision encoder and LLM, respectively. The projectors $\mathcal{P}_S$ and $\mathcal{P}_P$ are initialized from the pre-trained MLP projector of InternVL2.5-8B \cite{chen2024expanding}, while other projectors and the quality regression module are randomly initialized.
All the face video frames are resized to $448\times 448$ for model input.
The LoRA \cite{hu2022lora} rank is set to 8 for both vision encoder and LLM.
We train 2 epochs for each of the two stages (stage I and stage II) with a batch size of 8. The whole training process takes about 1 day on 2 NVIDIA A6000 GPUs (48G). Please refer to the \textit{Sup. Mat.} for more details.

\noindent\textbf{Baseline Methods.}
% Since there is no method specialized for the face video quality assessment (FVQA) task, we investigate and benchmark the state-of-the-art VQA, FIQA, facial beauty prediction (FBP), and LMM methods on the FVQA task, and compare the proposed FVQ-Rater with these methods.
Due to the lack of methods specifically tailored for the FVQA task, we systematically benchmark state-of-the-art methods from VQA, FIQA, Facial Beauty Prediction (FBP), and LMMs on the FVQA task, and conduct a comprehensive comparison with our proposed FVQ-Rater.
% The performance of the proposed FVQ-Rater is also evaluated by comparing with these methods:
% (1) general-purpose VQA methods: RAPIQUE \cite{tu2021rapique}, VIDEVAL \cite{tu2021ugc}, VSFA \cite{li2019quality}, GSTVQA \cite{chen2021learning}, SimpleVQA \cite{sun2022deep}, FastVQA \cite{wu2022fast}, Dover \cite{wu2023exploring}, and KSVQE \cite{lu2024kvq}; (2) FIQA methods: SDD-FIQA \cite{ou2021sdd} and DSL-FIQA \cite{chen2024dsl}; (3) facial beauty prediction (FBP) methods: REX-INCEP \cite{bougourzi2022deep}, 2D\_FAP \cite{liu2025lightweight}, and MEBeauty \cite{lebedeva2022mebeauty}; (4) LMM-based methods: Video-LLaVA \cite{lin2023video}, VideoLLaMA2 \cite{cheng2024videollama}, VideoLLaMA3 \cite{zhang2025videollama}, Qwen2-VL \cite{wang2024qwen2}, Qwen2.5-VL \cite{bai2025qwen2}, Deepseek-VL \cite{lu2024deepseek}, InternVL2.5 \cite{chen2024expanding}, and Q-Align \cite{wu2023q}.

\noindent\textbf{Evaluation Metrics.}
We use Spearman's Rank Correlation coefficient (SRCC), Pearson Linear Correlation Coefficient (PLCC), and Kendall's Rank Correlation Coefficient (KRCC) to evaluate the consistency between predicted quality scores and ground-truth MOSs. In addition, accuracy is used to evaluate the quality level prediction.

\subsection{Performance Evaluation}

\noindent\textbf{Comparison with state-of-the-art methods.}
We first benchmark the performance of state-of-the-art VQA methods, FIQA methods, and LMM models on the FVQA task as shown in Table \ref{tab:compare_mos}. We report the performance on the FVQ-20K dataset (including its TikTok \cite{tiktok} and YouTube \cite{youtube} subsets) and the CFVQA \cite{li2024perceptual} dataset.
It can be observed that pre-trained LMMs and most VQA methods perform poorly on the FVQA task, except for very recent methods such as Dover \cite{wu2023exploring}, KSVQE \cite{lu2024kvq} and DSL-FIQA \cite{chen2024dsl}, which achieve acceptable performance. 
% Note that the poor performance of DSL-FIQA \cite{chen2024dsl} on the CFVQA \cite{li2024perceptual} dataset may due to the fact that its degradation extraction module is pre-trained on the in-the-wild dataset and cannot generalize well to the compression scenario.
The proposed FVQ-Rater method achieves the best performance across all the datasets.
These results indicate that considering both temporal information and face-specific features can further improve the performance of FVQA compared to considering only temporal information or face-specific features.
It is worth mentioning that our FVQ-Rater is the first no-reference method specialized for the FVQA task, which even outperforms the existing state-of-the-art full-reference FVQA method FAVOR \cite{li2024perceptual} as shown in the last two rows of Table \ref{tab:compare_mos}.
Furthermore, both the superior performance of our FVQ-Rater and the promising performance of the fine-tuned LMMs highlight the great potential of LMMs in the quality assessment tasks.

Considering that in some practical scenarios, we only need approximate quality levels rather than precise quality scores of face videos, we also report the quality level classification accuracy of the state-of-the-art LMM-based methods and the proposed FVQ-Rater method in Table \ref{tab:compare_level}. Our FVQ-Rater achieves the highest quality level accuracy across all the subsets of FVQ-20K dataset and the CFVQA \cite{li2024perceptual} dataset, demonstrating its superior and robust performance not only in quality score regression but also in quality level prediction.

% \subsection{Cross-dataset Evaluation}
\noindent\textbf{Cross-dataset Evaluation.}
To evaluate the generalization ability of the proposed FVQ-Rater method, we also conduct two cross-dataset evaluations: (1) training on the training set of FVQ-20K and testing on the test set of CFVQA \cite{li2024perceptual}, and (2) training on the training set of CFVQA \cite{li2024perceptual} and testing on the test set of FVQ-20K. As shown in Table \ref{tab:cross}, our FVQ-Rater method achieves superior performance in both experimental settings, highlighting the strong generalization ability of our method.

\subsection{Ablation Study}
To validate the effectiveness of each component of our FVQ-Rater method, we conduct ablation studies on both FVQ-20K and CFVQA datasets as shown in Table \ref{tab:ablation}.

\noindent\textbf{Effectiveness of LoRA Fine-tuning.}
We first conduct ablation studies to evaluate the effectiveness of using LoRA \cite{hu2022lora} to fine-tune the pre-trained LLM and ViT (vision encoder).
As shown in the first three rows of Table \ref{tab:ablation}, compared to the baseline model that only trains the projector $\mathcal{P}_S$, LoRA fine-tuning of both LLM and ViT improves the performance of our model.

\noindent\textbf{Effectiveness of Quality-aware Pre-training.}
Then, we validate the effectiveness of the Stage I in our two-stage training framework, \textit{i.e.}, pre-training the projector using texts that describe the quality level of face videos. 
As shown in the third and fourth rows of Table \ref{tab:ablation}, quality-aware pre-training helps the model better understand the quality assessment task, thus improving its performance in quality score regression.

\noindent\textbf{Effectiveness of Multi-dimensional Features.}
Finally, we demonstrate the effectiveness of the proposed face-specific features and temporal features. Note that spatial features are necessary for the quality assessment task, and thus are included in all of our experiments.
The last four rows of Table \ref{tab:ablation} show that the performance of our model improves with the addition of each feature, which highlights the importance of introducing temporal features and face-specific features to the FVQA task.

\section{Conclusion}
In this paper, we explore the in-the-wild FVQA problem for the first time. 
Specifically, we collect the first large-scale FVQA dataset, FVQ-20K, which contains 20,000 face videos with MOS annotations.
Along with the FVQ-20K dataset, we propose the first LMM-based FVQA method, FVQ-Rater, which takes advantage of multi-dimensional features including spatial features, temporal features, portrait features, and achieves quality-related rating and scoring through the LoRA-based instruction tuning technique.
Extensive experiments on both FVQ-20K and CFVQA datasets demonstrate the superiority of our FVQ-Rater method.
We hope that the proposed FVQ-20K dataset, FVQA benchmarks, and FVQ-Rater method will promote in-depth research works on face video quality assessment.

%%
%% The acknowledgments section is defined using the "acks" environment
%% (and NOT an unnumbered section). This ensures the proper
%% identification of the section in the article metadata, and the
%% consistent spelling of the heading.
\begin{acks}
This work was supported in part by the National Natural Science Foundation of China under Grants 62401365, 62225112, 62132006, U24A20220, and in part by the China Postdoctoral Science Foundation under Grant Number BX20250411, 2025M773473.
\end{acks}

%%
%% The next two lines define the bibliography style to be used, and
%% the bibliography file.
\bibliographystyle{ACM-Reference-Format}
% \balance
\bibliography{sample-base}

%%-----------------------------------------------
%% Supplementary Material %%
%%-----------------------------------------------

\appendix
\clearpage
\renewcommand{\thetable}{\Alph{table}}
\renewcommand{\thefigure}{\Alph{figure}}
\setcounter{table}{0}
\setcounter{figure}{0}

\twocolumn[
\begin{center}
    \huge\bfseries Supplementary Material
    \vspace{2mm}
\end{center}
]

In the supplementary document, we provide more details about the proposed FVQ-20K dataset and FVQ-Rater method in Section \ref{sec:dataset} and Section \ref{sec:method}, respectively.

\section{More Details of the FVQ-20K Dataset}
\label{sec:dataset}

\begin{figure*}[h]
\centering
\includegraphics[width=\linewidth]{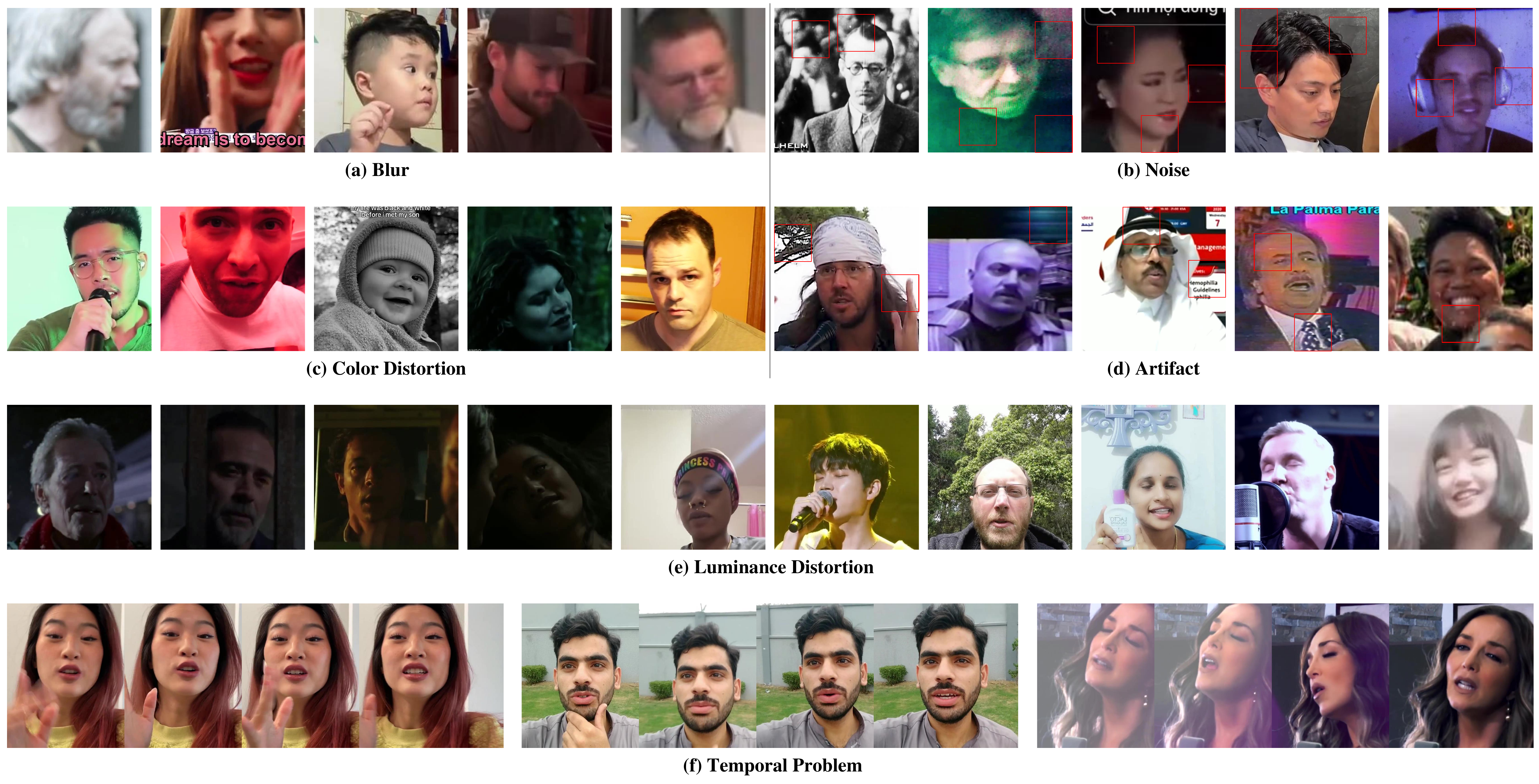}
\caption{Illustration of diverse in-the-wild distortions in the proposed FVQ-20K dataset, including (a) blur, (b) noise, (c) color distortion, (d) artifact, (e) luminance distortion, and (f) temporal problems such as motion blur, camera shake, and illumination variation. We encourage readers to zoom-in for details.}
\label{fig:demo_distortion}
\end{figure*}

\subsection{Face Video Collection}
\label{sec:video_collection}
The face videos in the FVQ-20K dataset are collected from two popular social media platforms, \textit{i.e.}, TikTok \cite{tiktok} and YouTube \cite{youtube}.

For the TikTok \cite{tiktok} part, we select 11 categories of video content where people appear most frequently, including relationship, beauty care, family, lipsync, society, shows, singing \& dancing, outfit, sports, comedy, and drama, and download a total of 100,212 videos from these categories.
Then, we filter out videos that do not contain faces and crop the face regions from the remaining videos to 5-second face videos through the commonly used face and landmark detection models \cite{zhang2016joint,bulat2017far}. 
Specifically, we use MTCNN \cite{zhang2016joint} to detect the bounding box of each facial region and add a margin around the bounding box to crop the original video into square face videos. If no bounding box is detected during the entire video, the video is filtered out. Otherwise, we use face-alignment \cite{bulat2017far} to detect 68 facial landmarks of each frame after the bounding box is detected to make sure the face is in the bounding box. If the detected facial landmark is out of bounding box and the obtained face video is less than 5 seconds, then we re-detect the bounding box and repeat the steps above until we accumulate a total of 5 seconds of face video or the original video ends. As a result, we obtain a total of 9,930 face videos from TikTok.

As for the YouTube \cite{youtube} part, since there are many publicly available portrait video datasets for other face-related tasks, we directly integrate and filter two recent datasets to construct the face video portion of our dataset.
Specifically, we merge the CelebV-HQ \cite{zhu2022celebv} dataset and the CelebV-Text \cite{yu2023celebv} dataset, and obtain a total of 104,218 face videos. Then we eliminate duplicate identities by simply retaining only one video from multiple videos that come from the same YouTube ID. Subsequently, we filter out videos shorter than 5 seconds and trim longer ones to 5 seconds to construct our final dataset. Eventually, we obtain a total of 10,070 face videos from YouTube.
In addition, we retrieve the category of each video based on its YouTube ID. The result shows that face videos from YouTube also span a variety of categories, including people \& blogs, entertainment, film \& animation, education \& technology, travel \& outdoor, music, news, sports, and comedy.

\begin{figure*}
\centering
\includegraphics[width=0.9\linewidth]{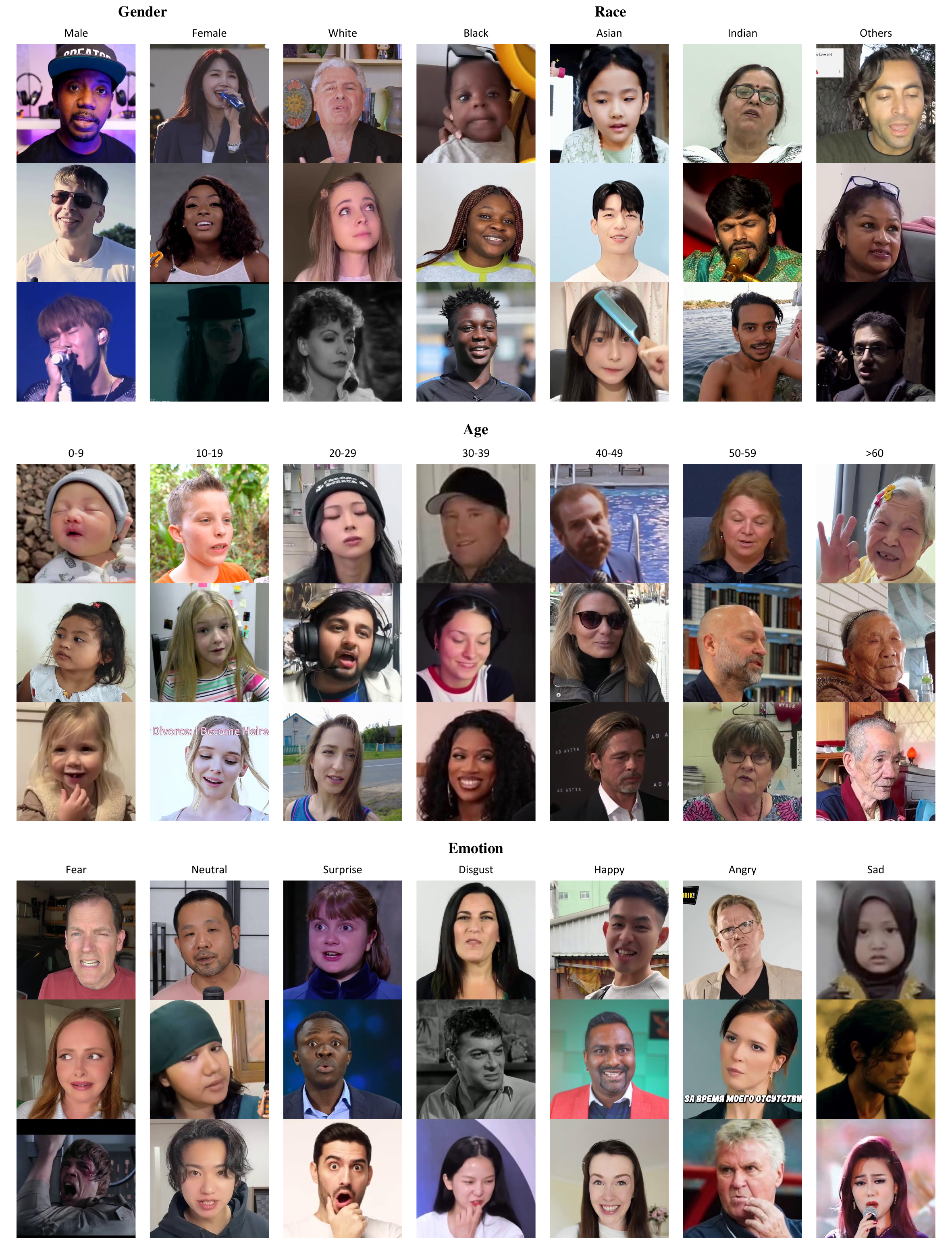}
\caption{Sample video frames from different genders, races, ages, and emotions in the proposed FVQ-20K dataset. We encourage readers to zoom-in for details.}
\label{fig:demo_attribute}
\end{figure*}

\subsection{Subjective Study}
The subjective study is conducted following the recommendation of ITU-BT.500 \cite{series2012methodology}. A total of 35 subjects participate in the experiment.

Considering the large scale of our dataset, we divide the videos into small batches to conduct the experiment.
Concretely, all 20,000 face videos are randomly shuffled and divided into 40 batches, with each batch containing 500 videos. Each video is played directly with the consistent video player to avoid the quality impact of manual decoding. Each subject is required to take a break after completing a batch of experiment, and each subject is limited to a maximum of two batches per half day to avoid inaccurate scoring caused by subject's excessive fatigue.

For each video, subjects are asked to rate the perceptual quality on a scale of 0 to 5, where the intervals 0–1, 1–2, 2–3, 3–4, and 4–5 correspond to bad, poor, fair, good, and excellent, respectively.
Before the formal experiments, we train all the subjects by explaining the detailed rating criteria (see Table \ref{tab:standard}) and numerous examples from each quality level. 
After the training session, we conduct a testing session with 15 face videos to evaluate whether the subjects are sufficiently trained. Only the subjects who pass the testing session are allowed to take part in the formal rating experiments. During the rating process, after completing each batch of data, we perform the subject rejection procedure \cite{series2012methodology} on each batch and exclude any subject identified as an outlier from participating in subsequent experiments. 

\begin{figure}
\centering
\includegraphics[width=\linewidth]{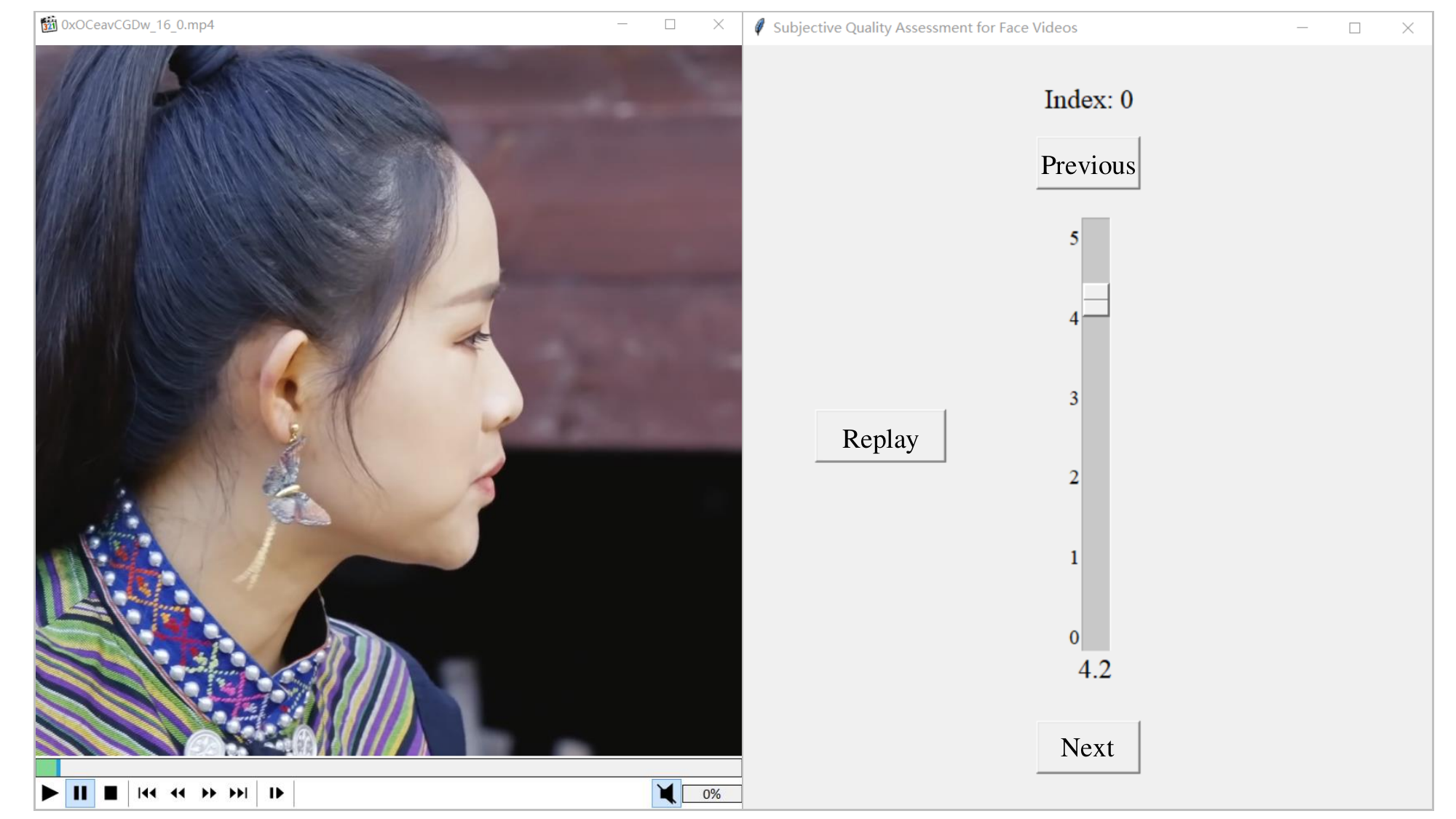}
\caption{Illustration of the GUI used in the subjective study. Videos are played in full screen during the experiment. The resolution of the slider is 0.1.}
\label{fig:gui}
\end{figure}

The subjective experiments are conducted on the graphical user interface (GUI) shown in Figure \ref{fig:gui}. The face videos are played in full screen. And the scoring interface only appears after each video has finished playing. The subjects are asked to use a slider to rate the video from 0 to 5, with 0 representing the worst and 5 representing the best. The resolution of the slider is 0.1. The subjects can freely replay or pause the video for details.

\begin{table*}
\centering
\caption{The rating criteria of the subjective study.}
\resizebox{\linewidth}{!}{
\begin{tabularx}{\textwidth}{>{\raggedright\arraybackslash}m{5cm} X}
\toprule
\textbf{Quality score range (quality level)}  & \textbf{Rating criteria} \\
\midrule
4-5 (Excellent)  & The video quality is excellent, presenting a clear and well-defined portrait. The lighting is natural and balanced. The camera and human motions are stable. \\
\midrule
3-4 (Good)  & The video quality is good, which is slightly worse or has a few minor problems compared to the ``excellent" one. For example, the video exhibits slightly reduced clarity and minor shakiness, however, the overall quality remains very high. \\
\midrule
2-3 (Fair)  & The video quality is moderate, with minor issues such as mild blurriness, shakiness, and unnatural colors. Nonetheless, the overall quality remains acceptable. \\
\midrule
1-2 (Poor)  & The video quality is relatively poor, with noticeable issues including significant color distortion, blurriness, low lighting conditions, and camera instability. \\
\midrule
0-1 (Bad)  & The video quality is extremely poor, exhibiting severe distortion that significantly impacts the overall visual presentation. \\
\bottomrule
\end{tabularx}
}
\label{tab:standard}
\end{table*}

\subsection{Subjective Data Processing}
To ensure the quality of the data, we first conduct outlier detection and subject rejection before converting the raw scores to Z-scores.
Concretely, we first calculate the kurtosis of the raw score for each video to determine whether the distribution is Gaussian or non-Gaussian. For the Gaussian cases, a raw score is considered an outlier if it lies beyond 2 standard deviations from the mean score of the corresponding video. For the non-Gaussian cases, a score is regarded as an outlier if it falls outside $\sqrt{20}$ standard deviations from the mean score. Finally, subjects with more than 5\% outlier scores are excluded.

\begin{figure}
\centering
\includegraphics[width=0.85\linewidth]{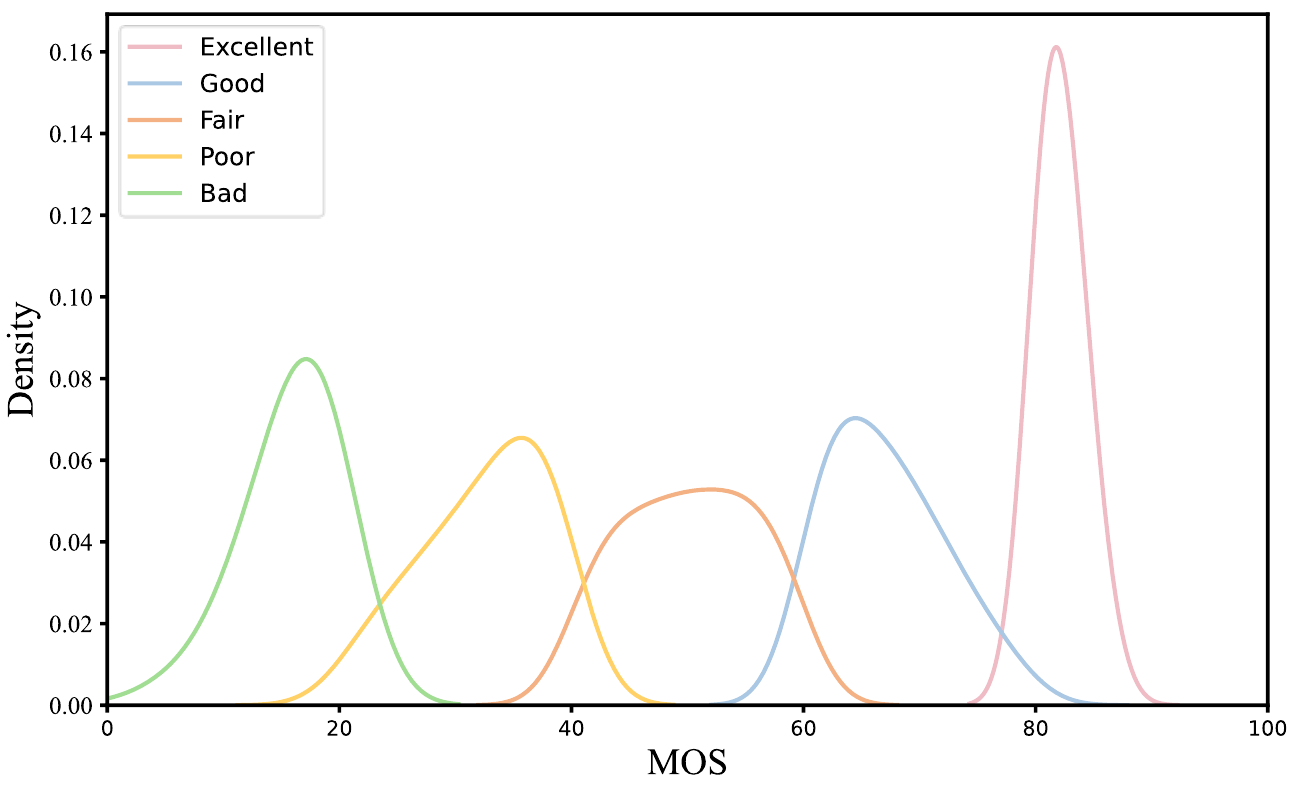}
\caption{The MOS distribution under different quality levels including excellent, good, fair, poor, and bad.}
\label{fig:mos_level}
\end{figure}

\begin{figure*}[h]
\centering
\includegraphics[width=\linewidth]{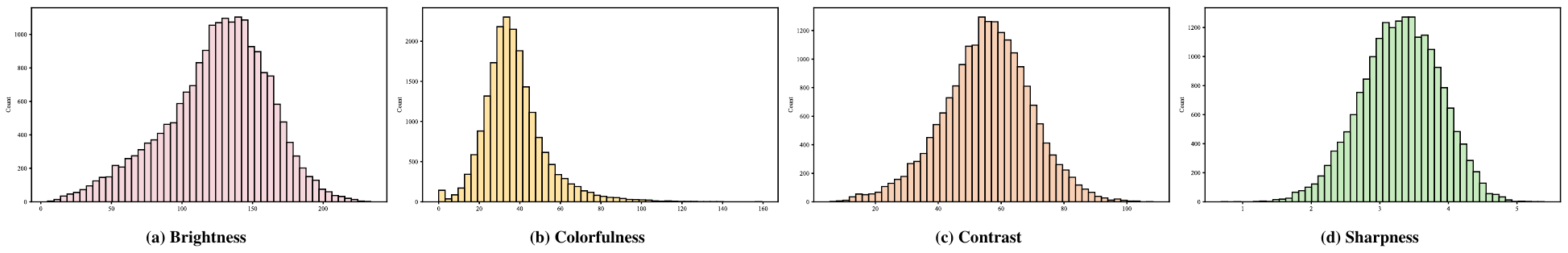}
\caption{The low-level feature distributions of the proposed FVQ-20K dataset, including (a) brightness, (b) colorfulness, (c) contrast, and (d) sharpness.}
\label{fig:brightness}
\end{figure*}

\subsection{Data Analysis}
We first show the variety of in-the-wild distortions present in the proposed FVQ-20K dataset in Figure \ref{fig:demo_distortion}. It can be observed that there are various distortions existed in the in-the-wild face videos, which demonstrates the significance of exploring the face video quality assessment (FVQA) problem.

We then show more demo frames of the face videos in the proposed FVQ-20K dataset, covering a variety of genders, races, ages, and emotions. As shown in Figure \ref{fig:demo_attribute}, our dataset contains a large variety of face videos with diverse face attributes.

Next, we plot the distribution of the brightness, colorfulness, contrast, and sharpness features of the face videos in our FVQ-20K dataset, as shown in Figure \ref{fig:brightness}. It can be observed that all the four low-level features span a wide range of values, indicating the inherent feature diversity of our dataset.
Specifically, we calculate the features for each video frame and then average them to obtain the features of the whole video. The brightness is estimated by the mean intensity of the V channel in the HSV color space. Contrast is measured as the standard deviation of pixel intensities in the grayscale version of the video. Colorfulness is computed based on the differences between the red (R), green (G), and blue (B) channels, which can be formulated as:
\begin{equation}
    \text{Colorfulness} = \sqrt{\sigma_{\text{rg}}^2 + \sigma_{\text{yb}}^2} + 0.3 \times \sqrt{\mu_{\text{rg}}^2 + \mu_{\text{yb}}^2},
\end{equation}
where $\sigma_{\text{rg}}$, $\sigma_{\text{yb}}$, $\mu_{\text{rg}}$, $\mu_{\text{yb}}$ denote the standard deviations and means of the $\text{rg}$ and $\text{yb}$ components, respectively. Specifically, $\text{rg}$ and $\text{yb}$ represent red-green (rg) and yellow-blue (yb) differences:
\begin{equation}
    \text{rg} = |R - G|, \quad \text{yb} = \left|0.5 \times (R + G) - B\right|.
\end{equation}
Sharpness is computed based on the gradient magnitude of the grayscale image using the Sobel operator, which can be formulated as:
\begin{equation} 
\text{Sharpness} = \log\left(1 + \text{mean}\left(\sqrt{G_x^2 + G_y^2}\right)\right),
\end{equation}
where $G_x$ and $G_y$ are the image gradients along the x and y directions, respectively.

Moreover, we plot the MOS distributions under the five quality levels, including excellent, good, fair, poor, and bad, as shown in Figure \ref{fig:mos_level}. It can be observed that the face videos in our FVQ-20K dataset cover a diverse perceptual quality range, and the MOS distribution under each quality level exhibits characteristics close to a normal distribution.

\begin{figure*}
\centering
\includegraphics[width=\linewidth]{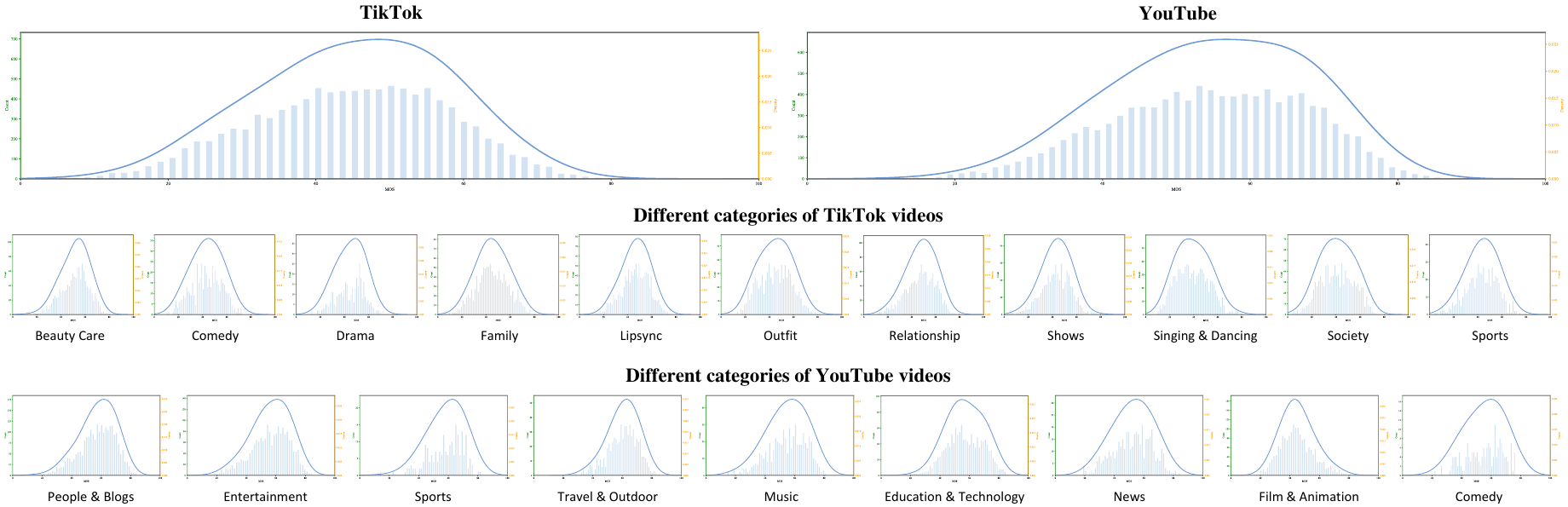}
\caption{The MOS distribution histograms and kernel density curves under different social media platforms (\textit{i.e.}, TikTok and YouTube) and corresponding video content categories.}
\label{fig:mos_platform}
\end{figure*}

\begin{figure*}
\centering
\includegraphics[width=\linewidth]{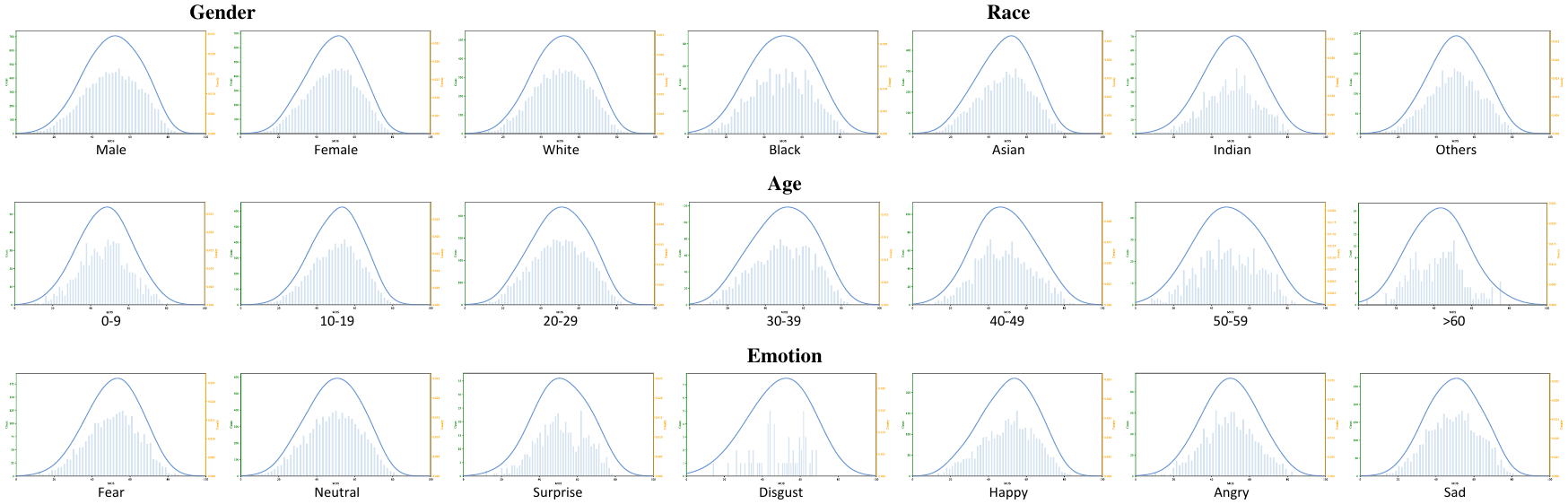}
\caption{The MOS distribution histograms and kernel density curves under different face attributes, including gender, race, age, and emotion.}
\label{fig:mos_attribute}
\end{figure*}

In addition, as a supplement to Figure 4 and Figure 5 in the main paper, we also plot the histograms and kernel density curves of the MOS distributions under different video content sources and face attributes in Figure \ref{fig:mos_platform} and Figure \ref{fig:mos_attribute}.
In Figure \ref{fig:mos_platform}, it can be observed that the MOS distribution of face videos from YouTube \cite{youtube} is kind of different from that of TikTok \cite{tiktok} videos, and the overall quality of face videos from YouTube is higher than those from TikTok, which indicates that videos from different platforms (such as short-form video platform TikTok and traditional video platform YouTube) exhibit distinct inherent distortions and quality problems. The MOS distributions are similar but vary across different video content categories. For example, the beauty care and lipsync contents in TikTok videos exhibit higher overall quality, whereas the singing \& dancing videos show lower quality due to the motion blur caused by dancing. The traditional video platform YouTube exhibits a higher quality compared to the short-form video platform, especially the travel \& outdoor content. By incorporating content from both types of video platforms, our FVQ-20K dataset achieves a more comprehensive coverage of in-the-wild face videos.

\section{More Details of the FVQ-Rater Method}
\label{sec:method}

\subsection{Network Architecture}
The extracted spatial, temporal, and face-specific features are projected into the LLM input embedding space via four distinct projectors, enabling their effective injection into the pre-trained LLM. The projectors are built upon the 2-layer multi-Layer perceptron (MLP), with each contains a layer normalization, a fully connected layer, a GELU \cite{hendrycks2016gaussian} activation function, and a fully connected layer in sequence. The input and output feature dimensions of the projectors depend on the extracted feature dimension and the LLM input feature dimension. Specifically, the input dimensions of the projectors $\mathcal{P}_S$, $\mathcal{P}_T$, $\mathcal{P}_P$, and $\mathcal{P}_F$ are 4096, 2304, 4096, and 128, respectively. And the LLM input feature dimension is 4096.

The quality regression module is built upon a 5-layer MLP with ReLU activation after each layer. The MLP takes 4096-dimensional features as input, and outputs 1-dimensional quality scores. The output dimensions of the fully connection layers in the MLP are 1024, 256, 64, 16, and 1, respectively.

\subsection{Loss Functions}
We utilize the cross entropy loss to supervise the text token prediction in the quality-aware pre-training stage (stage I), which can be formulated as:
\begin{equation}
    \mathcal{L}_{stageI} = - \frac{1}{N} \sum_{i=1}^{N} \log \left( \frac{\exp(z_{i, y_i})}{\sum_{k=1}^{C} \exp(z_{i, k})} \right),
\end{equation}
where $z_{i, c}$ is the logit for the $c$-th class of the $i$-th sample, $y_i$ is the ground truth class index, and $N$ is the number of samples in one training batch.

In the MOS-oriented LoRA fine-tuning stage (stage II), we use the L1 loss between the predicted quality score $\hat{s}_i$ and the ground truth $s_i$ MOS to supervise the training process, which can be formulated as:
\begin{equation}
\mathcal{L}_{stageII} = \frac{1}{N} \sum_{i=1}^{N} \left| \hat{s}_i - s_i \right|,
\end{equation}
where $N$ denotes the number of samples in one training batch.

\subsection{Implementation Details}
The proposed FVQ-Rater is trained on 2 NVIDIA RTX A6000 GPUs (48G) and flash-attention \cite{dao2022flashattention,dao2023flashattention2} is used to save the GPU memory.
We train our model using the AdamW optimizer \cite{loshchilov2017decoupled} with $\beta=(0.9,0.999)$. The learning rate is initially set to 1e-6, gradually increased to 4e-5 for warming up, and then gradually decreased to 2e-9 throughout the training process.

\end{document}